\documentclass[lettersize,journal]{IEEEtran}
\usepackage{amsmath,amsfonts}
\usepackage{algorithmic}
\usepackage{algorithm}
\usepackage{array}
\usepackage[caption=false,font=normalsize,labelfont=sf,textfont=sf]{subfig}
\usepackage{textcomp}
\usepackage{stfloats}
\usepackage{url}
\usepackage{verbatim}
\usepackage{graphicx}
\usepackage{cite}
\usepackage{amsmath}
\usepackage{amsthm}
\usepackage{booktabs}
\usepackage{algorithm}
\usepackage{algorithmic}
\usepackage[switch]{lineno}
\usepackage{amssymb}
\usepackage{multirow}
\usepackage[normalem]{ulem}
\useunder{\uline}{\ul}{}
\usepackage{amsfonts} 
\usepackage{fdsymbol}
\usepackage{enumitem}
\usepackage{color}
\newtheorem{myDef}{\textbf{Definition}}
\usepackage[hidelinks]{hyperref}
\begin{document}

\title{Graph Out-of-Distribution Generalization with Controllable Data Augmentation}


\author{{Bin Lu, Xiaoying Gan$^\dag$,~\IEEEmembership{Member,~IEEE,} Ze Zhao, Shiyu Liang, Luoyi Fu,~\IEEEmembership{Member,~IEEE,} Xinbing Wang,~\IEEEmembership{Senior Member,~IEEE,} Chenghu Zhou}
\thanks{\IEEEcompsocthanksitem Bin Lu, Xiaoying Gan, Ze Zhao and Xinbing Wang are with the Department of Electronic Engineering, Shanghai Jiao Tong University, China (e-mail: robinlu1209@sjtu.edu.cn, ganxiaoying@sjtu.edu.cn, zhaoze@sjtu.edu.cn, xwang8@sjtu.edu.cn)}
\thanks{Shiyu Liang is with John Hopcroft Center for Computer Science, Shanghai Jiao Tong University, China (e-mail: lsy18602808513@sjtu.edu.cn)}
\thanks{Luoyi Fu is with the Department of Computer Science and Engineering, Shanghai Jiao Tong University, China (e-mail: yiluofu@sjtu.edu.cn)}
\thanks{Chenghu Zhou is with the Institute of Geographical Science and Natural Resources Research, Chinese Academy of Sciences, China (e-mail: zhouch@lreis.ac.cn)}
\thanks{$^\dag$Xiaoying Gan is the corresponding author.}}

\markboth{Journal of \LaTeX\ Class Files,~Vol.~14, No.~8, August~2021}%
{Shell \MakeLowercase{\textit{et al.}}: A Sample Article Using IEEEtran.cls for IEEE Journals}


\maketitle

\begin{abstract}
Graph Neural Network (GNN) has demonstrated extraordinary performance in classifying graph properties.
However, due to the selection bias of training and testing data (e.g., training on small graphs and testing on large graphs, or training on dense graphs and testing on sparse graphs), distribution deviation is widespread. 
More importantly, we often observe \emph{hybrid structure distribution shift} of both scale and density, despite of one-sided biased data partition. 
The spurious correlations over hybrid distribution deviation degrade the performance of previous GNN methods and show large instability among different datasets. 
To alleviate this problem, we propose \texttt{OOD-GMixup} to jointly manipulate the training distribution with \emph{controllable data augmentation} in metric space.
Specifically, we first extract the graph rationales to eliminate the spurious correlations due to irrelevant information. Secondly, we generate virtual samples with perturbation on graph rationale representation domain to obtain potential OOD training samples. 
Finally, we propose OOD calibration to measure the distribution deviation of virtual samples by leveraging Extreme Value Theory, and further actively control the training distribution by emphasizing the impact of virtual OOD samples.
Extensive studies on several real-world datasets on graph classification demonstrate the superiority of our proposed method over state-of-the-art baselines. 
\end{abstract}


\section{Introduction}
\IEEEPARstart{G}{raph} Neural Networks (GNN) are powerful techniques to describe the attributes and relations of nodes, as well as the properties of whole graphs. Classifying the underlying labels of graphs is a fundamental problem with applications across many fields.
However, GNN suffers poor domain generalization when exposed to data with out-of-distribution (OOD) shift in testing. With the widespread application of graph classification in high-stake fields such as biomedicine and financial risk management~\cite{DBLP:conf/iclr/SunHV020,DBLP:conf/aaai/Hassani22,DBLP:conf/ijcai/WuLL0022}, the potential risks posed by distribution shifts make graph OOD generalization become a practical and non-trivial problem. 

\begin{figure}
    \centering
    \includegraphics[width=\linewidth]{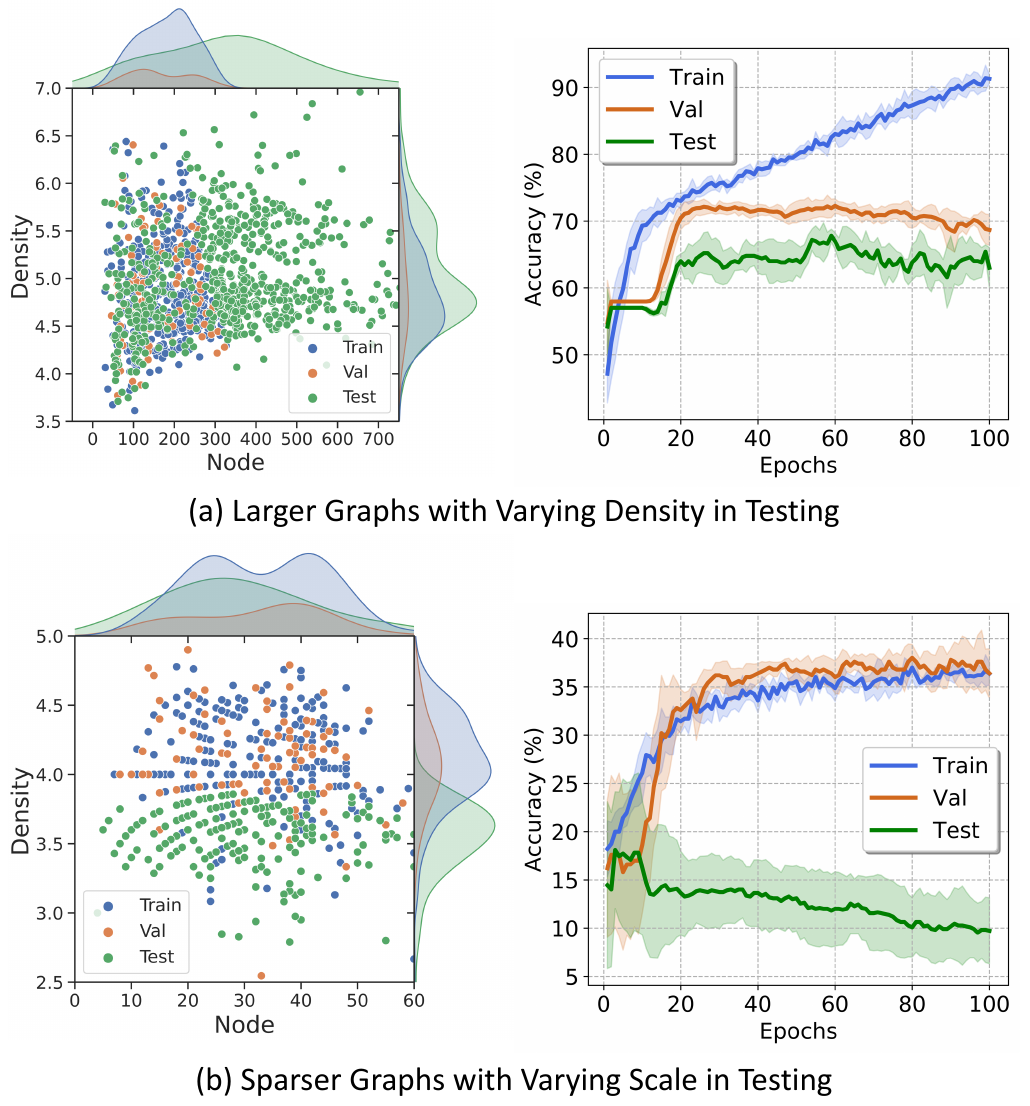}
    \caption{(a) Data partition via the number of nodes. (b) Data partition via the density of graphs. The hybrid structure distribution shifts of graphs make classical statistical learning paradigms, e.g., Empirical Risk Minimization (ERM), unable to be generalized to graph OOD scenarios. }
    \label{fig:intro}
\end{figure}

Compared with Euclidean data (e.g., images), graph distribution deviation exhibits more complex characteristics due to the graph structure.
We often observe \emph{hybrid structure distribution shift} that both \emph{scale shift} and \emph{density shift} occur simultaneously.
Figure \ref{fig:intro} depicts the structure distribution (left) and learning process (right) over two set of graph classification datasets with one-sided data partition. 
In Figure \ref{fig:intro}(a), although the selection bias is induced by scale (training on small graphs and testing on large graphs), the distribution of graph still shift in both scale and density manner.
Similarly, in Figure \ref{fig:intro}(b), data is partitioned only in density manner (training on dense graphs and testing on sparse graphs), while the scale of graphs shows obvious distribution deviation.
Due to the ignorance of this hybrid distribution shift, GNN with Empirical Risk Minimization (ERM) shows significant performance degradation and instability during testing.

Recently, there are some pioneering works on graph OOD generalization. 
On one hand, some previous work gain inspiration from the distribution generalization of images.
These works~\cite{OOD-GNN,GNN-DVD} consider the causality and invariance of node feature embeddings, but overlook the important role of graph structure.
On the other hand, some methods focus on \emph{one-sided} structure distribution shift, such as eliminating non-causal subgraphs through structure decomposition~\cite{DIR-GNN,chen2022invariance}, or propose regularization strategies for size generalization~\cite{buffelli2022sizeshiftreg}.
However, these methods show large instability across different graphs due to lack of calibration for training data with hybrid distribution deviation.
To manipulate the training distribution, data augmentation is an effective method to enlarge the span of training samples. However, directly adopting data augmentation approaches for hybrid graph distribution deviation still face two key challenges: (i) \emph{Single perturbation of data augmentation fails to generate diverse virtual samples.} Graph distribution shift is complex and even multiple ones. Existing graph data augmentation methods enhance the data from either pre-defined structural perturbation (e.g., edge perturbation, node dropping~\cite{virtual_node, dropedge, Dropnode}) or node feature perturbation (e.g., feature corruption, feature shuffling~\cite{DBLP:conf/www/WangWLCH21, flag}), which can not effectively adapt to complicated graph OOD scenarios.
(ii) \emph{Random perturbation of data augmentation fails to measure the representation deviation between ID and OOD distributions.} Randomized perturbation leads to unknown fluctuation of training distribution. Thus, the learning procedure cannot be effectively evaluated and actively controlled.

To address the aforementioned problems, we propose \texttt{OOD-GMixup} to jointly perform \emph{controllable data augmentation} in metric space for OOD generalization, which consists of three modules: graph rationale extraction, virtual sample generation, and out-of-distribution calibration. Firstly, in order to alleviate the model learning shortcuts from confounding factors, we identify the rationale of graphs by capturing the task-relevant patterns in both structure and feature. Secondly, instead of pre-defined single disturbance, we generate virtual samples with manifold mixup on graph rationale representations, so as to mimic the hybrid graph distribution shifts by disturbing representations in metric space. Thirdly, we propose OOD confidence score to measure the distribution deviation of virtual samples through Extreme Value Theory (EVT). A sample reweighting mechanism is further put forward to actively control the training procedure, distinguishing and strengthening the emphasis on virtual OOD samples. 
To summarize, the main contributions are as follows:
\begin{itemize}
    \item We propose \texttt{OOD-GMixup} to perform \emph{controllable data augmentation} to jointly manipulate the training distribution in metric space, providing a novel perspective to deal with hybrid structure distribution shift by investigating training samples.
    \item To our best knowledge, we are the first to explore Extreme Value Theory in OOD generalization. With the help of EVT, we propose OOD confidence score to calibrate the virtual training distribution, reducing the distribution deviation between ID and OOD graph representations.
    \item Extensive experiments on 6 real-world graph classification benchmarks with hybrid distribution shift demonstrate the superiority of our proposed methods over state-of-the-art baselines.
\end{itemize}

The rest of the paper is organized as follows. We start by reviewing related work in the next section. We then present the preliminary in Section \ref{sec:pre} and the details of our proposed \texttt{OOD-GMixup} in Section \ref{sec:method}. Experimental setup and discussion of results are provided in Section \ref{sec:experiment}. Finally, we conclude the paper and give some directions for future work in the last section.

\section{Related Work}
\label{sec:related_work}

In this section, we discuss the relevant research closely related to our work: graph out-of-distribution generalization, graph data augmentation and extreme value theory.

\subsection{Graph Out-of-Distribution Generalization}
In recent years, out-of-distribution generalization on graph has attracted increasing attention. Existing works can be categorized into three types: (1) \emph{improving the expressive power of GNN}: 
To enhance the graph representation, DisenGCN~\cite{DisenGCN} and FactorGCN~\cite{FactorGCN} aim to learn disentangled representation, which demonstrates to be more resilient to complex variants. 
TopKpool~\cite{TopK}, SAGpool~\cite{SAG} and PNA~\cite{PNA} improve pooling method to enhance the extraction of graph features. However, these works achieve satisfied results under i.i.d assumption, but fail to generalize in out-of-distribution scenarios. (2) \emph{invariant learning}: these approaches consider to learn domain invariant representations across different environments. Invariant Risk Minimization (IRM)~\cite{IRM} integrates variance over different domains in training to mitigate the over-reliance on data bias. Moreover, GroupDRO~\cite{GroupDRO} intend to regularize the worst-group cases. Whereas, invariant learning methods requires the annotation of different environment, which is high-cost and even infeasible for graph due to its hybrid distribution shift. (3) \emph{causality-based methods}: DIR-GNN~\cite{DIR-GNN} combines causal intervention to generate graph distributional perturbations, but the separation of graphs may destroy the connectivity of graphs. Others suggest cofounder balancing theory in causal inference. StableNet~\cite{StableNet} proposes sample reweighting via Hilbert-Schmidt Independence Criterion (HSIC) to eliminate correlation between features. OOD-GNN ~\cite{OOD-GNN} and GNN-DVD ~\cite{GNN-DVD} apply the similar idea to GNN. 
GNN-DVD explores graph generalization in node-level, and propose a decorrelation regularizer to eliminate the spurious correlation among labeled nodes. 
However, these approaches can only cope with graph feature shift or have large instability.

\subsection{Graph Data Augmentation}

Data augmentation is an effective method to improve the quantity and quality of training data. 
It is theoretically proved to improve the generalization by playing a regularization effect~\cite{DBLP:conf/icml/WuZVR20,DBLP:conf/iclr/ZhangDKG021}.
Existing graph data augmentation methods are mainly applied to graph self-supervised learning, especially graph contrastive learning, to generate different views~\cite{DBLP:conf/aaai/MoPXS022,DBLP:conf/nips/YouCSCWS20}. 
Ding et al.~\cite{DBLP:journals/sigkdd/DingXTL22} recently review representative graph data augmentation techniques, which can be mainly divided into structure-oriented, label-orientated and feature-oriented augmentation. 
For instance, DropEdge~\cite{dropedge} randomly removes a certain number of edges from the input graph at each training epoch. FLAG~\cite{flag} adversarially conduct feature corruption to generate various challenging samples. VirtualNode~\cite{virtual_node} connects all existing nodes to enhance long-distance message propagation.
However, existing graph data augmentation mainly rely on one-sided and random perturbations to generate virtual samples.
The effective control of data generation for hybrid distribution shift deserves further exploration.

\subsection{Extreme Value Theory}
Extreme Value Theory (EVT) is firstly proposed in statistical mathematics dealing with the extreme deviations from the median of probability distributions~\cite{de2006extreme}, which is further applied to a series optimization problems in disaster prediction, risk assessment, and human biology~\cite{einmahl2011ultimate,tippett2016more,songchitruksa2006extreme}. 

In recent years, some researchers combine EVT with deep learning problems, such as robustness analysis~\cite{robustness-analysis}, clustering~\cite{clustering}, and open set recognition~\cite{PMOSR,Openmax,C2AE}. 
Specifically, Weng et al.~\cite{robustness-analysis} convert robustness analysis into a local Lipschitz constant estimation problem, thereby utilizing EVT for estimation. Zheng et al.~\cite{clustering} propose the concept of centroid margin distance for clustering and use EVT to describe its distribution. As for open set recognition problem, Scheirer et al.~\cite{libMR} take the lead in applying EVT, and develop a new statistical predictor based upon the Weibull distribution. OpenMax~\cite{Openmax} use EVT to evaluate the open set risk and modify the classification probability. C2AE~\cite{C2AE} follow the Picklands-Balkema-deHaan formulation, and propose to use EVT to model the reconstruction error distribution. 
To the best of our knowledge, we are the first to apply 
Extreme Value Theory to OOD generalization.
With the help of EVT, we provide a new perspective to measure the distribution deviations of generated samples, and further reweight these virtual training samples to improve the generalization performance.

\section{Preliminary}
\label{sec:pre}

In this section, we first go over the notations used in this paper, and then introduce the problem definition of graph classification and graph out-of-distribution generalization. Then we provide a brief introduction of manifold mixup and extreme value theory, which will be used in our methodology design.

\subsection{Notations}

Given an input graph $\mathcal{G}(\mathcal{V}, \mathcal{E}, \mathbf{X})$, $\mathcal{V}$ and $\mathcal{E}$ denotes the node set $\{v_1, v_2, \cdots, v_n \}$ and edge set $\{e_1, e_2, \cdots, e_m\}$ representing the topology of graph. $\mathbf{X} = \left[ \mathbf{x_1}; \mathbf{x_2}; \cdots, \mathbf{x_n} \right] \in \mathbb{R}^{n \times d}$ is the node feature matrix, and each node $v \in \mathcal{V}$ is associated with a feature vector $\mathbf{x_{v}} \in \mathbb{R}^{1 \times d}$.
More generally, the attributed network can also be represented as $\mathcal{G}(\mathbf{A}, \mathbf{X})$.
$\mathbf{A}=\{a_{ij}\}^{n \times n}\in \mathbb{R}^{n \times n}$ is the adjacency matrix of graph. $a_{ij}=1$ indicates that there is an edge between node $v_i$ and $v_j$; otherwise, $a_{ij}=0$. Meanwhile, we summarize the main notations used throughout the paper in Table \ref{tab:symbol}. For the other additional notations, we will illustrate them in the corresponding section.

\begin{table}[h]
\centering
\caption{Table of main symbols}
\label{tab:symbol}
\resizebox{0.95\linewidth}{!}{%
\begin{tabular}{@{}cl@{}}
\toprule
\textbf{Symbols} & \textbf{Definitions} \\ 
\midrule
$\mathcal{G}$ & input attributed graph \\
$\mathcal{V}$ & Node set of input graph \\
$\mathcal{E}$ & Edge set of input graph \\
$\mathbf{X}$ & attribute matrix \\
$\mathbf{A}$ & adjacency matrix \\ 
$\mathbf{M}$ & structure masking \\
$\eta$ & learnable feature masking \\
$\mathcal{G}^{r}$ & graph rationale of graph $\mathcal{G}$ \\
$\mathbf{p}_{k}$ & class prototype of class $k$ \\
$z_i$ & latent representation of graph $\mathcal{G}_i$ in metric space\\
$\tilde{z}$ & virtual graph representation via manifold mixup\\
$\lambda$ & manifold mixup ratio \\
$\omega$ & OOD confidence score\\
$\overline{\omega}$ & normalized OOD confidence score\\
\bottomrule
\end{tabular}
}
\end{table}

\subsection{Problem Formulation}

\begin{myDef}
\textbf{Graph Classification.} 
Given a set of graphs, graph classification task aims to assign a class label of the entire graph. The goal of graph classification is to learn a mapping function $f:\mathcal{G} \rightarrow y$, where $y$ is the target label or category associated with the entire graph.
\end{myDef}

\begin{myDef}
\textbf{Graph Out-of-Distribution Generalization.} Given a set of training graphs of $N$ instances $\mathcal{D}=\{(\mathcal{G}_i, y_i)\}_{i=1}^{N}$ that are drawn from the training distribution $\mathcal{P}_{\text{train}}(\mathcal{G}, Y)$. The goal of graph out-of-distribution generalization for graph classification is to learn an optimal graph predictor $f$ with parameter $\theta^{*}$ that can achieve the best generalization on the data drawn from an unknown testing distribution $\mathcal{P}_{\text{test}}$, where $\mathcal{P}_{\text{train}}(\mathcal{G}, Y) \neq \mathcal{P}_{\text{test}}(\mathcal{G}, Y)$.     
\begin{equation}
    \theta^{*} = \text{arg}\min_{\theta} \mathbb{E}_{(\mathcal{G}_i,y_i)\sim \mathcal{P}_{\text{test}}} \mathcal{L}(f_{\theta}(\mathcal{G}_i), y_i).
\end{equation}
\end{myDef}

Unlike domain adaptation, we are not exposed to a specific target domain during training while generalize to multiple unseen domains, which makes it more challenging but more realistic in practice.

\subsection{Manifold Mixup}
Mixup~\cite{mixup} is a simple but effective data augmentation technique to construct virtual training samples by linear interpolations of raw data as follows:
\begin{equation}
    \tilde{x} = \lambda x_{i} + (1 - \lambda) x_{j}, \; \tilde{y} = \lambda y_{i} + (1 - \lambda) y_{j},
\end{equation}
where $(x_{i}, y_{i})$ and $(x_{j}, y_{j})$ are two samples randomly selected from training data, $\lambda \in [0,1]$ is sampled from a Beta distribution \texttt{Beta}$(\alpha,\beta)$.
However, due to the irregularity in the structure and scale of graph data, mixup cannot be directly conducted, so we adopt the idea of manifold mixup~\cite{DBLP:conf/icml/VermaLBNMLB19,DBLP:conf/www/WangWLCH21} to generate virtual graph representations $\tilde{z}$:
\begin{equation}
    \tilde{z} = \lambda \phi(\mathcal{G}_i) + (1-\lambda) \phi(\mathcal{G}_j),  \;   \tilde{y} = \lambda y_{i} + (1 - \lambda) y_{j}
\end{equation}
where $(\mathcal{G}_i, y_i)$ and $(\mathcal{G}_j, y_j)$ are two graph samples, $\phi(\cdot): \mathcal{G} \rightarrow \mathcal{Z}$ is the GNN backbone for graph representation. In our work, we only conduct manifold mixup on graph representations within the same label. We hope to generate a new data distribution $\hat{\mathcal{P}}$ from the training data distribution $\mathcal{P}_{\text{train}}$, so that we can actively generate various OOD data in training, thereby improving the generalization performance.

\subsection{Extreme Value Theory}
Extreme value theory is a statistical branch that studies the behavior of extreme events, which model their occurrence probabilities and estimate extreme quantiles in datasets.
In this paper, we follow the Picklands-Balkema-deHaan formulation of the extreme value theory~\cite{de2006extreme}. It considers modeling probabilities conditioned on random variable exceeding a high threshold. Given a sequence of independent and identically distributed random variables $X_1, X_2, \cdots, X_n$ with cumulative distribution function (CDF) $F(x)$. Assuming that there is a sufficiently large threshold $u$, $X_i - u$ is the excess, then the CDF of the excess $F_u$ is
\begin{equation}
    F_{u}(x)=P(X-u \leqslant x \mid X>u)=\frac{F(x+u)-F(u)}{1-F(u)},
\end{equation}
where $x \geq 0$. Further, given a large enough $u$, $F_u$ can be well approximated by the Generalized Pareto Distribution:
\begin{equation}
    G(x)=1-\left(1+\xi \frac{x-\mu}{\sigma}\right)^{-1 / \xi},
\end{equation}
where $x \geq \mu$, $1+\xi \frac{x-\mu}{\sigma}>0$. $\mu, \sigma, \xi$ are location, scale and shape parameter respectively, and $\mu \in R, \sigma>0, \xi \in R$.
In Section \ref{sec:ood_calibration}, guided by EVT, we propose how to calibrate distribution deviation of virtual samples and further improve OOD generalization.

\begin{figure*}
    \centering
    \includegraphics[width=\linewidth]{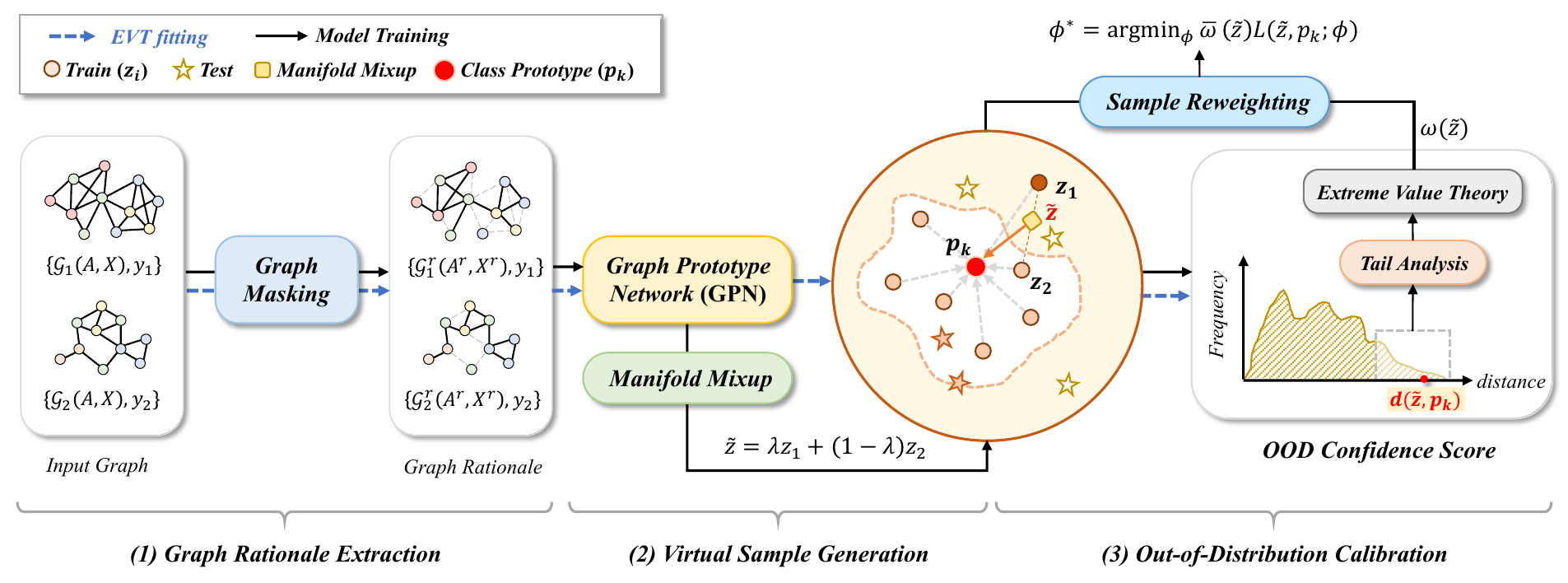}
    \caption{The framework of proposed $\texttt{OOD-GMixup}$, which consists of the following three steps: (1) \emph{Graph Rationale Extraction}: graph masking retrieves the rationale part $\mathcal{G}^r$ from the input graph $\mathcal{G}$ to eliminate the spurious correlations due to irrelevant graph information. (2) \emph{Virtual Sample Generation}: Virtual training samples are generated with manifold mixup on two graph representations with the same label. (3) \emph{Out-of-Distribution Calibration}: Extreme value theory is utilized to establish a probability model to measure distribution deviation for sample reweighting.}
    \label{fig:system_model}
\end{figure*}

\section{Methodology}
\label{sec:method}

In this section, we elaborate \texttt{OOD-GMixup} as shown in Figure \ref{fig:system_model}. The main idea of our model is to investigate the rationale of graphs, generate and distinguish efficient virtual OOD training samples through controllable augmentation, so as to continuously improve OOD generalization.


\subsection{Graph Rationale Extraction}
\label{sec:graph_rationale}

Existing post-hoc GNN explainability studies~\cite{DBLP:conf/kdd/SuiWWL0C22,DBLP:journals/corr/abs-2204-11028} typically build an explainer model to decompose the input graphs according to their importance and sample the salient features as an explanatory subgraph. 
The spurious structure information (e.g. noisy links, non-causal subgraphs) induces biased representation and further lead to the failure of OOD generalization. Hence, we first propose graph masking to discover the graph rationale $\mathcal{G}^r$ for the input graph $\mathcal{G}$, which consists of structure masking and feature masking. 

We first generate a soft structure mask $\mathbf{M} \in \mathbb{R}^{|V| \times |V|}$ according to the semantic correlation of two adjacent nodes on $\mathbf{A}$, where $\mathbf{M}_{ij}$ indicates the importance of edge $a_{ij}$. 
\begin{equation}
    \mathbf{H} = \mathbf{NN}(\mathbf{X}), \quad \mathbf{M}_{ij} = \sigma{(\mathbf{H}_i^T \mathbf{H}_j)},
\end{equation}
where $\mathbf{H} \in \mathbb{R}^{|V| \times p}$ represents the $p$-dimensional representations of all nodes. $\mathbf{NN}(\cdot)$ denotes a neural network and we use a one-layer perceptron. $\sigma(\cdot)$ is the sigmoid function, which project the edge importance into the range of $(0,1)$.
A weighted adjacency matrix of graph rationale is obtained by $\mathbf{A}^r = \mathbf{M} \odot \mathbf{A}$. 
Then, we use a learnable feature masking $\eta \in \mathbb{R}^{d}$ to drop features that are irrelevant to the downstream task, which is formulate as $X^r = \eta \odot X$. Intuitively, if a particular feature is an irrelevant factor, the corresponding weight in $\eta$ takes value close to zero. 

According to above two soft masking, we derive the graph rationale $\mathcal{G}^r(\mathbf{A}^r, \mathbf{X}^r)$ of each input graph, which eliminates the confounding shortcuts caused by spurious graph structure. The identification of graph rationale enables task-relevant structural learning, thereby obtaining more accurate graph representations.

\subsection{Virtual Sample Generation}
\label{sec:virtual}
Secondly, after the graph rationale is obtained, we discuss how to effectively generate virtual training samples that reflects different graph distribution shifts. 
\paragraph{Graph Prototype Network (GPN)}
Prototype network is a classification method by measuring the feature distance with each class prototype~\cite{DBLP:conf/nips/SnellSZ17,DBLP:conf/kdd/Geometer} in metric space. 
Compared with classical classifier, prototype network classifies by pairwise sample comparison, which is more robust to the distribution deviation of representations.
In our work, we adopt this idea to transform the graph classification problem into the distance measurement between sample representations and class prototypes.
The graph rationale of $\mathcal{G}_i$ is encoded into representation ${z}_i \in \mathbb{R}^{M}$ through an embedding function $\phi(\cdot): \mathcal{G}^r \rightarrow z$. Generally, $\phi(\cdot)$ is a multi-layer GNN followed by a pooling layer (e.g., mean pooling, max pooling, etc.), or other graph-level representation learning methods. Each class prototype $\{\mathbf{p}_1, \cdots, \mathbf{p}_K\}$ is $K$ mean vector of the graph embedding belonging to its class.
\begin{equation}
    \label{eq:pn}
    \mathbf{p}_{k} = \frac{1}{N_k} \sum_{\{\mathcal{G}_{i}^{r}, y_{i}=k\}_{i=1}^{N_k}} \phi(\mathcal{G}_{i}^{r}),
\end{equation}
where $N_k$ is the number of graphs with label $y=k$ in training data.
Furthermore, a distance function $d(\cdot, \cdot)$ is defined to calculate the distance between graph representation and class prototype. Commonly, squared Euclidean distance is a simple and effective way. Therefore, the prediction probability over label can be defined as the average negative log-likelihood probability of true class $y=k$:
\begin{equation}
    p_{\phi}(\hat{y}_i=k|z_i) = \frac{\exp{-d(z_i,\mathbf{p}_k)}}{\sum_{k^{\prime}} \exp{-d(z_i, \mathbf{p}_{k^{\prime}})}}.
\end{equation}

\paragraph{Manifold Mixup}
Due to the complex graph distribution shift, single perturbation of graph augmentation methods fail to express diverse distribution deviation. Therefore, we consider the disturbance from graph representation level.
According to Vicinal Risk Minimization (VRM) principle~\cite{vrm}, we propose to sample graphs within same label from training distribution $\mathcal{P}_{\text{train}}$, and use manifold mixup~\cite{DBLP:conf/icml/VermaLBNMLB19,DBLP:conf/www/WangWLCH21} to obtain virtual graph representations through linear interpolation.
\begin{equation}
    \tilde{z} = \lambda z_i + (1-\lambda) z_j,  \;   \tilde{y} = \lambda y_{i} + (1 - \lambda) y_{j},
\end{equation}
where $(z_i, y_i)$ and $(z_j, y_j)$ are two graph representation pairs with same label $y_i=y_j$. $\lambda \in [0,1]$ is sampled from a Beta distribution \texttt{Beta}$(\alpha,\beta)$. Therefore, the generated graph representation distribution $\hat{\mathcal{P}}$ is approximated by
\begin{gather}
    \label{eq:vrm}
    \hat{\mathcal{P}}(\tilde{z},\tilde{y}) = \frac{1}{n} \sum_{i=1}^{n} \mu (\tilde{z}, \tilde{y} | z_i, y_i), \\
    \mu (\tilde{z}, \tilde{y} | z_i, y_i) = \frac{1}{N_i}\sum_{y_i = y_j} \mathbb{E}_{\lambda} [\delta(\tilde{z}=\lambda z_i + (1-\lambda) z_j, \tilde{y}=y_i)],
\end{gather}
where $N_k$ is the sample number of class $k$, $\delta(\tilde{z}, \tilde{y})$ is a Dirac mass centered at $(\tilde{z}, \tilde{y})$.
We hope to actively strengthen virtual OOD samples from the generated graph representation distribution $\hat{\mathcal{P}}$, so that the model can gradually improve generalization performance during training. 
However, random manifold mixup cannot evaluate the distribution deviation, which leads valuable OOD samples to be hidden in the in-distribution virtual samples. 
Therefore, an effective measurement for the distribution deviation of virtual samples are desired for controllable augmentation.  

\subsection{Out-of-Distribution Calibration}
In this subsection, we discuss how to calibrate the distribution deviation of virtual samples and further enhance the adaptability of GNN to OOD samples.
\label{sec:ood_calibration}
\paragraph{OOD Confidence Score}
Since we use graph prototype network for classification, the distance between virtual mixup samples and class prototype indicates the distribution shift relative to the in-distribution samples. In particular, samples with a larger prototype distance in training indicate a larger distribution deviation and are more likely to be OOD data. Therefore, we model the distributions of distances to class prototype over a threshold by Generalized Pareto Distribution in Extreme Value Theory (EVT). Correspondingly, we establish a probability model for each class, and perform the evaluation of distribution deviation based on our proposed OOD confidence score.

\begin{algorithm}[htb]
\caption{EVT calibration for out-of-distribution samples, with per class Weibull fitting to $\tau$ largest distance between virtual graph embedding and class prototype.}
\label{alg:algorithm_evt_fitting}
\textbf{Input}: FitHigh function from \texttt{libMR}.\\
\textbf{Input}: For each class $k$, let $\mathcal{G}(k) = \{\mathcal{G}_i, y_i=k\}$ denote each correctly classified training sample.\\
\textbf{Output}: Class prototype $\mathbf{p}_k$ for each class $k$.\\
\textbf{Output}: \texttt{libMR} models $\rho_k$, which includes location $\mu$, scale $\sigma$ and shape $\xi$.
\begin{algorithmic}[1] 
\FOR{$k = 1, \cdots, n$}
\STATE Derive the graph rationale $\mathcal{G}^{r}$ of each graph;
\STATE Calculate class prototype $\mathbf{p}_k$ in Eq. \ref{eq:pn};
\STATE EVT Fit $\rho_k=\{\mu_k, \sigma_k, \xi_k\}$ = FitHigh($d(\mathcal{G}^{r}(k), \mathbf{p}_k)$, $\tau$);  // \emph{Maximum Likelihood Estimation}
\ENDFOR
\STATE \textbf{return} Class prototypes $\mathbf{p}_k$, and \texttt{libMR} models $\rho_k$.
\end{algorithmic}
\end{algorithm}

To be specific, we use the \texttt{libMR}~\cite{libMR} FitHigh function to do Weibull fitting on $\tau$ largest of the distances between all correct positive training samples and class prototype. The EVT model parameters $\rho_k$ per class are obtained as shown in Algorithm \ref{alg:algorithm_evt_fitting}. Then, we utilize the Weibull CDF probability on the distance between virtual graph representation and class prototype to estimate the probability of falling into out-of-distribution data, which is denoted as the OOD confidence score $\omega$. For example, the confidence score of manifold mixup sample $\tilde{z}_i$ with label $y=k$ can be calculated as follows:
\begin{equation}
    \label{eq:ood_prob}
    \omega(\tilde{z}_i) = 1 - \exp({-(\frac{d(\tilde{z}_i, p_k) - \mu_k}{\sigma_k})^{\xi_k}})
\end{equation}
where $\mu_k, \sigma_k, \xi_k$ are EVT model parameter of class $y=k$. 

\paragraph{Sample Reweighting}
Furthermore, we propose a sample reweighting strategy to enhance the learning of these OOD virtual graph representation. Our intuition is that a virtual sample with a larger OOD confidence score should contribute more to promote the OOD generalization. Therefore, instead of averaging the sample loss in each batch, we normalize the confidence score in each mini-batch of batch size $B$ as:
\begin{equation}
    \label{eq:sample_weight}
    \overline{\omega}(\tilde{z}_i) = \frac{\omega(\tilde{z}_i)}{\frac{1}{B}\sum_{j=1}^{B}\omega(\tilde{z}_j)}.
\end{equation}
The loss function is further reweighted as follows:
\begin{equation}
    \label{eq:loss}
    \mathcal{L}_{\texttt{OOD-GMixup}} = \sum_{(\tilde{z}_i,\tilde{y}_i) \sim \hat{\mathcal{P}}} \overline{\omega}(\tilde{z_i}) \cdot \log p_{\phi}(\hat{y}_i=\tilde{y_i}|\tilde{z_i}),
\end{equation}
where $\hat{y}_i$ is the prediction label given $\tilde{z}_i$. In order to explain the learning of \texttt{OOD-GMixup} more clearly, we summarize the training procedure in Algorithm \ref{alg:ood_mixup}.

\begin{algorithm}[tb]
\caption{The training procedure of $\texttt{OOD-GMixup}$}
\label{alg:ood_mixup}
\textbf{Input}: Training data $\mathcal{D}_{\text{train}} = \{\mathcal{G}, \mathbf{Y}\}$.\\
\textbf{Output}: \texttt{OOD-GMixup} model parameter.\\
\textbf{Initialization}: Initialize model parameter $\phi$ with random uniform distribution.
\begin{algorithmic}[1] 
\STATE Iteration $t \leftarrow 0$;
\WHILE{not converged or $t < \text{maxIter}$}
\FOR{$j = 1, \cdots, n$}
\STATE Calculate class prototype $p_j$ and fit EVT model parameter $\rho_j$ according to Algorithm \ref{alg:algorithm_evt_fitting};
\ENDFOR
\STATE Generate virtual graph representations distribution $\hat{\mathcal{P}}$ with graph manifold mixup from $\mathcal{P}_{\text{train}}$ via Eq. \ref{eq:vrm};
\FOR{$\tilde{z}$ in virtual training distribution $\hat{\mathcal{P}}$}
\STATE Calculate its OOD confidence score $\omega(\tilde{z})$ and sample weights $\overline{\omega}(\tilde{z})$ via Eq. \ref{eq:ood_prob} and Eq. \ref{eq:sample_weight};
\STATE Optimize GNN model parameter $\phi$ to minimize $\mathcal{L}_{\texttt{OOD-GMixup}}$ in Eq. \ref{eq:loss}.
\ENDFOR
\STATE $t = t + 1$;
\ENDWHILE
\STATE \textbf{return} \texttt{OOD-GMixup} model parameter $\phi^{*}$
\end{algorithmic}
\end{algorithm}

\section{Experiment}
\label{sec:experiment}

In this section, we evaluate the effectiveness of our proposed \texttt{OOD-GMixup} on six real-world datasets and conduct extensive ablation studies. More comprehensive in-depth analysis (including learning patterns and insight, hyperparameter sensitivity, etc.) are presented in detail with the aim of answering the following four research questions.
\begin{itemize}
    \item \textbf{RQ1}: How does $\texttt{OOD-GMixup}$ perform against other baselines in out-of-distribution generalization?
    \item \textbf{RQ2}: How effective is each part of the proposed method?
    \item \textbf{RQ3}: What are the learning patterns and insights of $\texttt{OOD-GMixup}$ during training?
    \item \textbf{RQ4}: How does each major hyperparameter affect the performance during training process?
\end{itemize}




\subsection{Datasets} 

We conducted experiments on 6 real-world datasets. We introduce the basic information, data division method and the hyperparameter setting of each dataset as follows. 
In Table \ref{tab:appendix_data}, we summarize the detailed statistics. 
It is worth noting that we select three different one-sided data partition method based on the number of nodes, number of edges, and graph density, and all datasets show hybrid distribution shifts, which is consistent with our observations.

\begin{itemize}
    \item \textbf{Movie collaboration datasets}: \emph{IMDB-BINARY} is a movie collaboration dataset where each graph represents a movie and the nodes denote actors/actresses, and an edge exists if they appear in the same movie. The task is to predict whether the movie belongs to the romance genre or the action genre. To simulate the selection bias, we choose 400/100 of the graphs with less than 20 nodes as training, validation, and the rest are divided into testing sets. \emph{IMDB-MULTI} is multi-class version of \emph{IMDB-BINARY} and contains a balanced set of ego-networks derived from Comedy, Romance and Sci-Fi genres. Similarly, we choose 600/150 of the graphs with more than 10 nodes as training, validation, and the rest are divided into testing sets.
    \item \textbf{Social Network datasets}: \emph{REDDIT-BINARY} corresponds to the discussion network in Reddit: nodes in each graph represent users, and edges exist if there is communication between users, and the graphs correspond to two types: question/answer-based community and discussion-based community. Based on the edge density, we select 1200/200 graphs with density less than 2.4 for training and validation, and the rest are used as the testing set. \emph{REDDIT-MULTI} is a larger variant of \emph{REDDIT-BINARY} from five different subreddits, namely, worldnews, videos, AdviceAnimals, aww and mildlyinteresting. We select 3000/500 graphs with density more than 2.1 for training and validation, and the rest are used as the testing set.
    \item \textbf{Scientific collaboration dataset}: \emph{COLLAB} graph represents the central network of researchers: researchers are nodes and edges represent the existence of collaboration between two people. The task is to divide the graph into the following three categories: High Energy Physics, Condensed Matter Physics, and Astro Physics, representing the domain to which the researchers belong. We select 1500/500 in graphs with less than 1000 edges as training and validation, and the rest as testing set.
    \item \textbf{Bioinformatics datasets}: \emph{D\&D} is a dataset containing 1178 protein structures. Each protein is represented as a graph where nodes are amino acids and two nodes are connected if the distance between them is less than 6$\mathring{A}$. The prediction task is to classify the structure of proteins into enzymes and non-enzymes. We spilt the dataset according to the number of edges per graph, and for graphs with less than a specific threshold number of edges 400/100 were selected for training and validation, and the rest for testing.    
\end{itemize}
All above datasets are public\footnote{https://pytorch-geometric.readthedocs.io/en/latest/modules/datasets.html}, and thanks to PyTorch Geometric (PyG) library for providing an easy-to-use method to load all the datasets.

\begin{table}[]
\centering
\caption{Detailed statistics of experiment datasets.}
\label{tab:appendix_data}
\resizebox{\linewidth}{!}{%
\begin{tabular}{@{}lllrrrr@{}}
\toprule
Dataset & Bias & Split & \multicolumn{1}{l}{\#Graphs} & \multicolumn{1}{l}{Avg.N} & \multicolumn{1}{l}{Avg.E} & \multicolumn{1}{l}{Avg.D} \\ \midrule
\multirow{3}{*}{IMDB-B} & \multirow{3}{*}{\#Node} & Train & 400 & 14.96 & 128.05 & 8.56 \\
 &  & Val & 100 & 15.12 & 121.22 & 8.02 \\
 &  & Test & 500 & 24.57 & 259.82 & 10.57 \\ \midrule
\multirow{3}{*}{IMDB-M} & \multirow{3}{*}{\#Node} & Train & 600 & 17.49 & 200.70 & 9.57 \\
 &  & Val & 150 & 18.31 & 230.31 & 9.80 \\
 &  & Test & 750 & 8.32 & 56.91 & 6.58 \\ \midrule
\multirow{3}{*}{REDDIT-B} & \multirow{3}{*}{Density} & Train & 1200 & 454.41 & 1014.59 & 2.23 \\
 &  & Val & 200 & 490.45 & 1083.81 & 2.21 \\
 &  & Test & 600 & 359.79 & 927.91 & 2.58 \\ \midrule
\multirow{3}{*}{REDDIT-M} & \multirow{3}{*}{Density} & Train & 3000 & 554.59 & 1318.76 & 2.31 \\
 &  & Val & 500 & 563.95 & 1330.54 & 2.30 \\
 &  & Test & 1500 & 397.81 & 884.60 & 2.11 \\ \midrule
\multirow{3}{*}{COLLAB} & \multirow{3}{*}{\#Edge} & Train & 1500 & 43.77 & 551.75 & 12.61 \\
 &  & Val & 500 & 43.44 & 564.71 & 13.00 \\
 &  & Test & 3000 & 95.03 & 7820.73 & 82.30 \\ \midrule
\multirow{3}{*}{D\&D} & \multirow{3}{*}{\#Edge} & Train & 400 & 180.11 & 875.26 & 4.86 \\
 &  & Val & 100 & 167.66 & 821.88 & 4.90 \\
 &  & Test & 678 & 363.00 & 1849.26 & 5.09 \\ \bottomrule
\end{tabular}%
}
\end{table}

\subsection{Implementation Details}

The number of epochs is set to 200. We train our model via the Adam optimizer with a learning rate of 0.001
By default, we use GCN as the feature extractor.
The dimension of graph representation is chosen from \{32, 64, 128\}. The num of GNN layers is chosen from \{1,2,3\}. We set the parameter of \texttt{Beta}$(\alpha, \beta)$ distribution in manifold mixup as $\alpha=2$ and $\beta \in \{1,2,3,4\}$.
All models are trained with early stopping strategy (i.e., monitoring the performance on validation dataset and stopping the training process when the model's performance starts to deteriorate). 

To support the reproducibility of the results, we have released our code\footnote{https://anonymous.4open.science/r/OOD-GMixup/}. We implement the $\texttt{OOD-GMixup}$ model based on Pytorch 1.8.1 framework and PyG (PyTorch Geometric 2.0.2) library. Weibull fitting functionality is implemented in LibMR 0.1.9 library. All the evaluated models are implemented on a server with two CPUs (Intel Xeon E5-2630 × 2) and four GPUs (NVIDIA GTX 2080 × 4, 12GB memory).

\subsection{Baselines}

In the experiment, we compared with ERM and other 16 baseline models. Here we introduce each baseline model in detail. In general, these baseline models can be divided into the following four categories.

(1) Improving the expressive power of GNN
\begin{itemize}
    \item \textbf{Attention}~\cite{GAT}: Calculate the node feature correlations with the attention mechanism for feature aggregation, which is a widely used GNN backbone in many applications.
    \item \textbf{TopKpool}~\cite{TopK}: Using a projection vector to transform nodes into corresponding scores, and only nodes with Top K scores along with related edges are remained for each pooling.
    \item \textbf{SAGPool}~\cite{SAG}: A hierarchical graph pooling method, which utilizes self-attention mechanism to calculate whether nodes should be deleted or retained for global pooling.
\end{itemize}

(2) Graph data augmentation
\begin{itemize}
    \item \textbf{Virtual Node}~\cite{virtual_node}: Add a virtual node connected with all the nodes in graph, so that it can contain the global information for better graph representation.
    \item \textbf{DropoutEdge}~\cite{dropedge}: Randomly remove a certain number of edges from the input graph at each training epoch.
    \item \textbf{DroputNode}~\cite{Dropnode}: Randomly remove a certain number of nodes from the input graph at each training epoch.
    \item\textbf{Mixup}~\cite{DBLP:conf/www/WangWLCH21}: Mix up graphons of different classes for data augmentation, realizing the topology interpolation of different graphs.
    \item \textbf{FLAG}~\cite{flag}: Augment the features of graph nodes by introducing adversarial perturbations in training.
\end{itemize}

(3) Domain Generalization
\begin{itemize}
    \item \textbf{IRM}~\cite{IRM}: Integrate the variance of different pre-defined domains in the training process to reduce the over-reliance on data bias.
    \item \textbf{Group DRO}~\cite{GroupDRO}: Minimize the worst-case training loss over a set of pre-defined groups. 
    \item \textbf{V-REx}~\cite{vrex}: Negative weighting is proposed to achieve interpolation between distributions, thus achieving better generalization performance under distribution bias.
    \item \textbf{IB-IRM}~\cite{ib-irm}: Combine information bottleneck theory with invariant learning to improve the generalization performance of the model.
    \item \textbf{CAD}~\cite{DBLP:conf/iclr/RuanDM22}: Provide minimal sufficient objectives whose optima achieve optimal DG under covariate shift that preserves the Bayes predictor.
\end{itemize}

(4) Graph OOD Generalization Methods
\begin{itemize}
    \item \textbf{DIR-GNN}~\cite{DIR-GNN}: DIR-GNN divides the graph into causal and non-causal subgraphs for causal intervention, so that generating distribution perturbations adaptively to remove spurious correlations.
    \item \textbf{OOD-GNN}~\cite{OOD-GNN}: OOD-GNN employs a novel nonlinear graph representation decorrelation method utilizing random Fourier features.
    \item \textbf{SizeShiftReg}~\cite{buffelli2022sizeshiftreg}: SizeShiftReg proposes a regularization strategy to solve size-shift problem which  generates coarsened graph and minimizes the distribution differences between coarsened embeddings with the original embeddings.
\end{itemize}

It should be noted that we do not compare GNN-DVD~\cite{GNN-DVD} and EERM~\cite{wu2022handling} as baselines, because both methods are aimed at node classification, and we pay more attention to out-of-distribution generalization in graph classification. In addition, the graphs we considered in the experiment all have node features, while G-Mixup~\cite{G-Mixup} only applies to undirected graphs without node features, and therefore is not within the scope of our baselines.

\begin{table*}[]
\centering
\caption{Performance comparison of graph OOD generalization on 6 datasets, where we abbreviate BINARY as B and MULTI as M. \textbf{bold} indicates the best results, and \underline{Underlined} indicates second-best results. The last column records the average ranking on each dataset. Our proposed \texttt{OOD-GMixup} outperforms other state-of-the-art baselines.}
\label{tab:performance}
\resizebox{0.9\linewidth}{!}{%
\begin{tabular}{@{}rrrrrrrc@{}}
\toprule
Baseline & \multicolumn{1}{c}{IMDB-B} & \multicolumn{1}{c}{IMDB-M} & \multicolumn{1}{c}{REDDIT-B} & \multicolumn{1}{c}{REDDIT-M} & \multicolumn{1}{c}{COLLAB} & \multicolumn{1}{c}{D\&D} & $\overline{\text{Rank}}$ \\ \midrule
ERM & 52.34±3.69 & 33.43±4.15 & 62.26±4.62 & 44.70±4.31 & 43.61±4.19 & 63.61±3.70 & 16 \\ \midrule
Attention~\cite{GAT} & 54.59±3.03 & 34.17±4.09 & 63.40±4.95 & 43.81±4.78 & 43.46±4.80 & 62.54±3.50 & 13 \\
Top-K Pooling~\cite{TopK} & 47.84±9.96 & 31.23±1.65 & 58.83±5.43 & 38.81±2.68 & 47.02±4.17 & 66.27±5.94 & 15 \\
SAGPooling~\cite{SAG} & 44.25±6.70 & 32.99±3.48 & 61.91±6.92 & 37.06±8.99 & 46.29±4.05 & 63.39±2.28 & 18 \\ \midrule
Virtual Node~\cite{virtual_node} & 45.21±5.49 & 34.91±3.60 & 62.43±1.56 & 28.81±0.97 & 46.45±4.66 & 56.64±1.59 & 17 \\
DropoutEdge~\cite{dropedge} & 52.32±3.13 & 33.55±2.67 & 68.66±1.04 & 28.77±9.15 & 31.26±4.68 & 60.28±1.65 & 14 \\
DropoutNode~\cite{Dropnode} & 54.23±4.92 & 35.27±3.14 & {\ul 70.17±4.12} & 27.59±8.59 & 43.42±4.27 & 61.46±8.20 & 12 \\
FLAG~\cite{flag} & 46.42±4.53 & 34.61±2.82 & 60.29±2.91 & 38.45±0.70 & {\ul 51.20±1.88} & 58.35±1.08 & 11 \\
Mixup~\cite{DBLP:conf/www/WangWLCH21} & 58.44±8.49 & 34.98±4.69 & 67.60±2.01 & {\ul 48.53±2.54} & 47.56±4.28 & 71.53±1.27 & 6 \\ \midrule
Group DRO~\cite{GroupDRO} & 56.12±4.32 & 34.37±2.56 & 30.17±5.77 & 39.17±3.17 & 30.17±5.77 & 71.74±1.13 & 10 \\
IRM~\cite{IRM} & 52.78±6.92 & 35.15±2.68 & 34.47±6.07 & 31.91±6.94 & 34.47±6.07 & 69.22±3.19 & 9 \\
V-REx~\cite{vrex} & 55.36±4.84 & 35.33±1.62 & 32.60±5.17 & 34.58±5.90 & 32.60±5.17 & 71.17±1.63 & 7 \\
IB-IRM~\cite{ib-irm} & 53.91±4.32 & 34.61±2.49 & 42.34±9.86 & 30.04±4.69 & 42.34±9.86 & 68.96±2.72 & 8 \\
CAD~\cite{DBLP:conf/iclr/RuanDM22} & 55.31±5.25 & 35.03±2.04 & 65.38±0.61 & 37.99±0.90 & 48.22±2.45 & 65.80±2.68 & 5 \\ \midrule
DIR-GNN~\cite{DIR-GNN} & 47.94±6.21 & 32.87±3.75 & 65.08±6.99 & 39.77±4.77 & 51.01±1.50 & 65.28±2.13 & 4 \\
OOD-GNN~\cite{OOD-GNN} & {\ul 58.64±3.30} & 20.27±1.57 & 64.10±2.58 & 28.53±0.74 & 49.51±3.46 & {\ul 74.42±2.22} & 3 \\
SizeShiftReg~\cite{buffelli2022sizeshiftreg} & 51.46±4.23 & {\ul 35.35±4.32} & 65.23±5.53 & 33.02±0.87 & 45.78±3.51 & 70.55±2.02 & 2 \\ \midrule
\texttt{OOD-Mixup} (Ours) & \textbf{66.79±3.59} & \textbf{38.40±4.00} & \textbf{70.55±2.08} & \textbf{50.46±1.41} & \textbf{57.50±1.81} & \textbf{79.41±0.57} & 1 \\ \bottomrule
\end{tabular}%
}
\end{table*}

\begin{figure*}[h]
    \centering
    \includegraphics[width=0.9\linewidth]{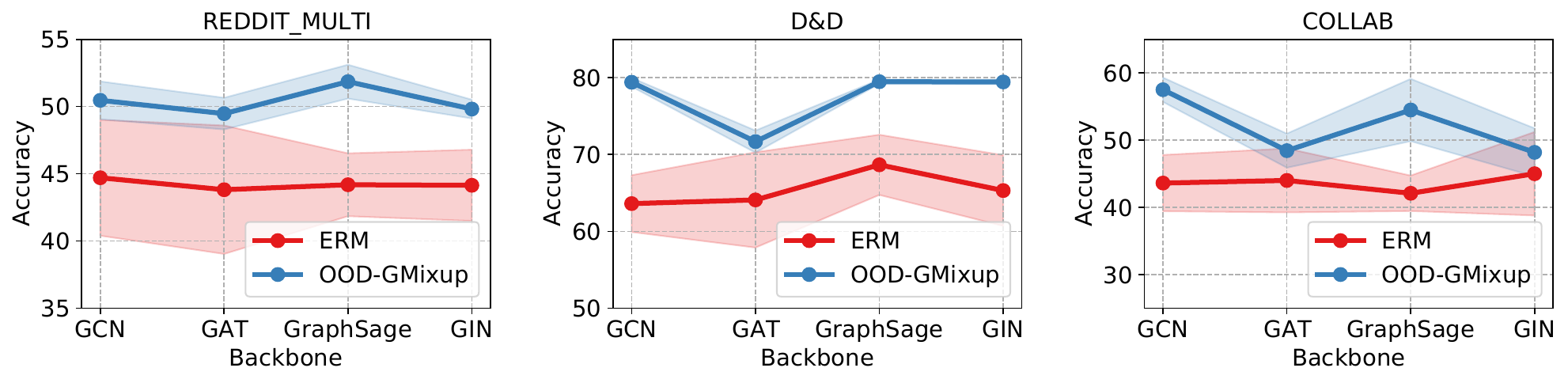}
    \caption{Performance Comparison on different GNN backbone: GCN, GAT, GraphSage, GIN. \texttt{OOD-GMixup} shows consistent improvement over ERM.}
    \label{fig:backbone}
\end{figure*}

\subsection{Performance Comparison (\textbf{RQ1})}

We run \texttt{OOD-GMixup} and other baselines more than 10 times with different random seeds and report the average test accuracy and standard deviation on different distribution shift in Table \ref{tab:performance}. We have the following observations: 

First, \texttt{OOD-GMixup} significantly outperforms other baselines on 6 datasets, and rank the first on the averaging performance over all datasets, demonstrating the superiority of our proposed method. Notably, for \emph{IMDB-BINARY}, \emph{COLLAB}, \emph{IMDB-MULTI} and \emph{D\&D} datasets, our model achieves 13.90\%, 12.30\%, 8.62\%, and 6.71\% improvements over state-of-the-art model respectively. This performance boost suggests that \texttt{OOD-GMixup} can achieve stable results over hybrid distribution shift.

Second, data augmentation shows its effectiveness, while the controllable strategy of \texttt{OOD-GMixup} further promote its stability. Some data augmentation methods (e.g., Mixup, DropoutNode, FLAG) show suboptimal performance on some datasets, and partially surpass many advanced OOD generalization algorithms, indicating that a suitable pre-defined data augmentation method can improve OOD generalization to some extent, which is neglected in previous studies. However, due to their single and random disturbance, these methods show large instability among different datasets.

Third, \texttt{OOD-GMixup} can deal with hybrid distribution shifts more effectively. 
Overall, three graph OOD generalization methods achieve 2nd to 4th place in average rankings. However, the performance of these methods fluctuates greatly on different datasets.
SizeShiftReg is specifically designed to generalize from small to large graph data, but ignores the influence of density. DIR-GNN intervenes the distribution through graph structure, while the separation of graphs may destroy the connectivity of graphs. OOD-GNN decorates the feature embedding, but the spurious structure information may lead to poor generalization. Our proposed \texttt{OOD-GMixup} attempts to generate diverse virtual samples and calibrate the OOD samples by EVT, demonstrating more stable performance.


On the other hand, we replace different GNN backbones, i.e., GCN, GAT, GraphSage and GIN on three representative datasets. As shown in Figure \ref{fig:backbone}, \texttt{OOD-GMixup} significantly improve performance compared to ERM, with smaller variance. The performance improvement is between 12.84.14\% - 17.36\% on REDDIT-MULTI dataset, 11.84\% - 24.84\% on D\&D dataset, and 7.07\% - 31.85\% on COLLAB dataset, which indicates the consistent enhancement of OOD generalization for different backbones.

\begin{figure*}
    \centering
    \includegraphics[width=0.9\linewidth]{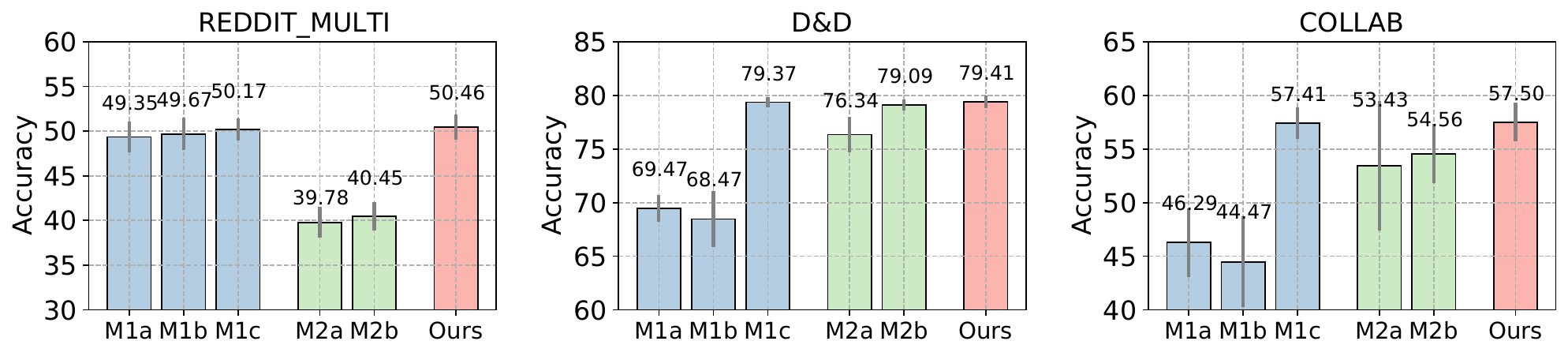}
    \caption{Ablation studies of \texttt{OOD-GMixup} with different variants on REDDIT-MULTI, D\&D and COLLAB dataset.}
    \label{fig:ablation}
\end{figure*}

\subsection{Ablation Studies (\textbf{RQ2})}
In order to verify the effectiveness of different modules in our proposed \texttt{OOD-GMixup}, we conduct ablation studies with following two groups of variants: \emph{(M1) Variants of Graph Rationale Extraction}, and \emph{(M2) Variants of OOD Calibration}. Due to space limitations, the ablation studies are conducted on three representative datasets, i.e. REDDIT-MULTI, D\&D, and COLLAB.

\begin{itemize}
    \item \textbf{(M1a) w/o Graph Masking}:  Removing graph masking (both feature and structure) strategy.
    \item \textbf{(M1b) w/o Structure Masking}: Removing graph structure masking, only retaining feature masking.
    \item \textbf{(M1c) w/o Feature Masking}: Removing graph feature masking, and only retaining structure masking.
    \item \textbf{(M2a) w/o EVT-based Sample Reweighting Loss}: Removing EVT-based OOD calibration, and use normalized prototype distance directly to calculate sample weights.
    \item \textbf{(M2b) Focal Loss}: Due to the imbalance of virtual training samples, using focal loss~\cite{focal_loss} instead of EVT-based method.
\end{itemize}

As shown in Figure \ref{fig:ablation}, first of all, due to the possibility of spurious correlation in both feature and structure, the best result is obtained when using joint graph masking. In particular, the structure rationale plays a more important role owing to the message-passing nature of GNN. Secondly, by comparing different learning strategy, \texttt{OOD-GMixup} is more effective than other variants. 
Compared with using prototype distance directly or entropy-based focal loss, distribution modeling based on extreme distance can depict out-of-distribution deviation better, and further reduce the representation deviation between ID and OOD distributions.

\begin{figure}[h]
    \centering
    \includegraphics[width=0.95\linewidth]{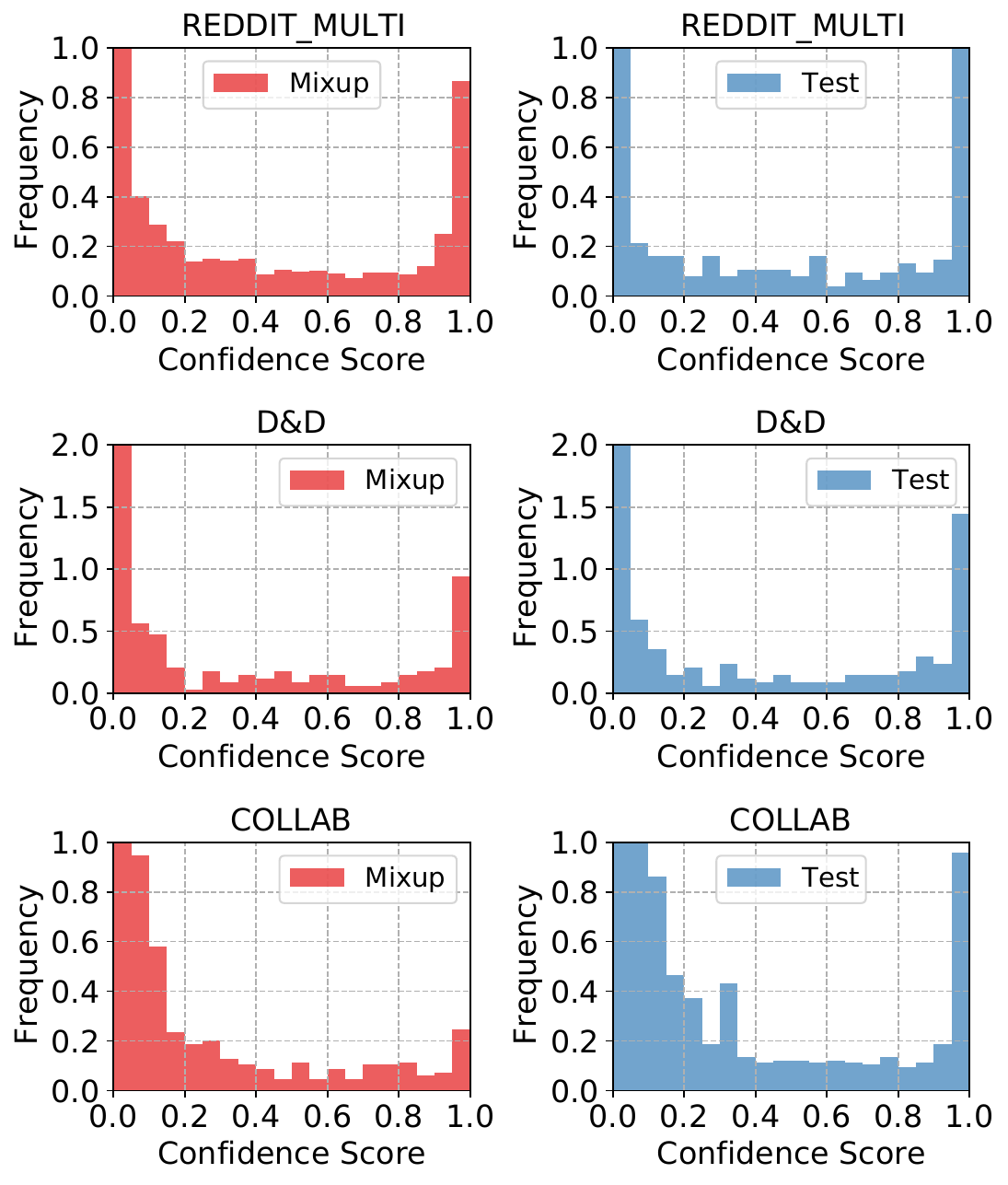}
    \caption{Comparison of confidence scores between manifold mixup samples and test data based on ERM. Mixup can effectively generate virtual graph representation under various distribution shift. Simultaneously, modeling the distance distribution based on EVT can effectively evaluate and detect out-of-distribution data.}
    \label{fig:score}
\end{figure}

\subsection{Model Analysis (\textbf{RQ3})}
In order to better analyze the learning patterns and insight of \texttt{OOD-GMixup}, we propose the following three questions for in-depth study.

\textbf{Whether manifold mixup can generate virtual out-of-distribution samples?}
We conduct the following experiments to validate that manifold mixup can generate valuable virtual graph representations. We train the GNN model based on ERM principle and calculate confidence scores on mixup samples and test data accordingly.
As depicted in Figure \ref{fig:score}, the mixup sample covers numerous confidence score intervals in test samples, suggesting that the mixup method may successfully widen the distribution of training data. This enables additional samples that are originally outside the training data distribution to join the training procedure and improves the OOD generalization ability. 

\begin{figure*}
    \centering
    \includegraphics[width=\linewidth]{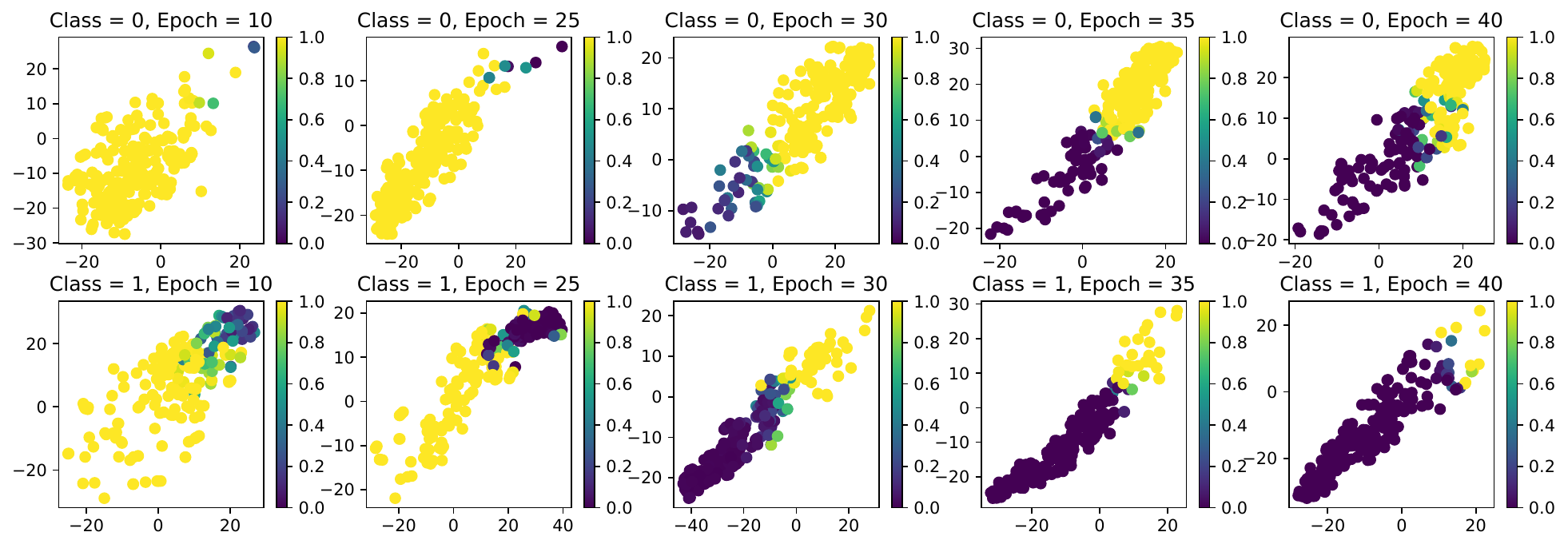}
    \caption{A $t$-SNE visualization of virtual training samples on D\&D dataset. The color indicates OOD confidence score of virtual samples.}
    \label{fig:tsne}
\end{figure*}

\begin{figure*}
    \centering
    \includegraphics[width=0.9\linewidth]{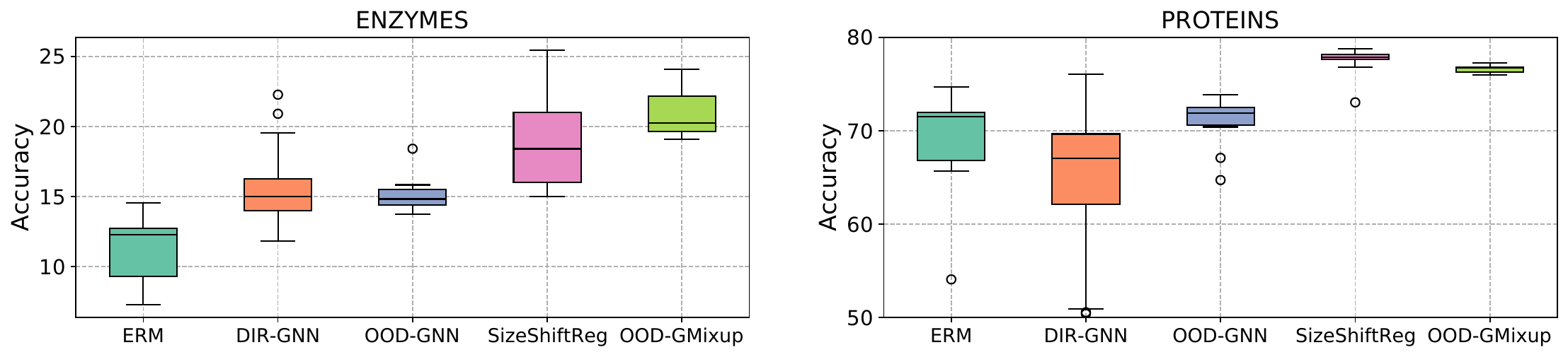}
    \caption{Performance Comparison on two one-sided distribution deviation datasets. Our proposed \texttt{OOD-GMixup} shows competitive performance.}
    \label{fig:oneside}
\end{figure*}

\textbf{Whether OOD confidence score can measure the adaptability of GNN to OOD samples?}
In order to further analyze the OOD confidence score of \texttt{OOD-GMixup}, we take D\&D dataset as an example, and visualize its training process as shown in Figure \ref{fig:tsne}. The element in each figure denotes the virtual graph representations, and the color represents the confidence score per epoch. When the score is closer to 0, it means that the mixup samples belong to the in-distribution data, and the adaptability of GNN is better, which can be better classified; On the contrary, when the score is closer to 1, it means that it tends to the out-of-distribution data, and the GNN model has poor performance. 
In the early stage of model training, the classification ability of mixup samples is poor due to the poor adaptability to OOD data. With the continuous training of the model, more and more virtual samples can be effectively identified (dark colored samples),  but there are still many OOD samples that are difficult to distinguish. When it comes to the 40th epoch, most of the mixup samples can be accurately classified, and the model gradually converges, realizing the adaptability to OOD samples.

\begin{table}[]
\centering
\caption{Statistics of two one-sided distribution shift datasets}
\label{tab:one_side}
\resizebox{0.95\linewidth}{!}{%
\begin{tabular}{@{}clcrrrr@{}}
\toprule
Dataset & Bias & spilt & \multicolumn{1}{c}{\#Graphs} & \multicolumn{1}{c}{Avg.N} & \multicolumn{1}{c}{Avg.E} & \multicolumn{1}{c}{Avg.D} \\ \midrule
\multirow{3}{*}{ENZYMES} & \multirow{3}{*}{Density} & Train & 280 & 33.12 & 133.97 & 4.05 \\
 &  & Val & 220 & 32.46 & 131.36 & 4.05 \\
 &  & Test & 220 & 32.09 & 108.71 & 3.39 \\ \midrule
\multirow{3}{*}{PROTEINS} & \multirow{3}{*}{\#Node} & Train & 400 & 15.35 & 57.14 & 3.72 \\
 &  & Val & 100 & 15.45 & 57.74 & 3.74 \\
 &  & Test & 613 & 58.44 & 217.94 & 3.73 \\
 \bottomrule
\end{tabular}%
}
\end{table}

\textbf{How does \texttt{OOD-GMixup} perform on graph datasets with \emph{one-sided} distribution deviation?} Although we observe a large amount of data with hybrid distribution shift, a few data do have one-sided distribution deviation, e.g. ENZYMES and PROTEINS.
In Table \ref{tab:one_side}, we illustrate the statistics information of two datasets. 
As shown in Figure \ref{fig:oneside}, we compare our proposed method with ERM and three latest graph OOD generalization methods. 
On \emph{ENZYMES} dataset, the distribution of graph density shifts but the scale remains consistent. Our proposed \texttt{OOD-GMixup} achieve the state-of-the-art performance among all baselines, which demonstrates the consistent advantage of hybrid distribution deviation. 
\emph{PROTEINS} dataset exhibits one-sided distribution deviation of graph scale, but the average density remains. SizeShiftReg~\cite{buffelli2022sizeshiftreg} achieves the best results with accuracy 77.42±1.54\%, which shows its capability in generalizing from small to large graph data through regularization.
Our proposed \texttt{OOD-GMixup} shows the competitive performance with accuracy 76.62±0.40\%, outperforming other graph OOD generalization baselines. 

\subsection{Hyperparameter Sensitivity (\textbf{RQ4})}

We investigate the hyperparameter sensitivity of \texttt{OOD- GMixup}, including the parameter $\beta$ in Beta distribution, tail size $\eta$ for EVT modeling and the layer of GNN.
We show the experiment results on REDDIT-MULTI, D\&D and COLLAB datasets, and other datasets have the similar patterns. 

\begin{figure}[h]
    \centering
    \includegraphics[width=\linewidth]{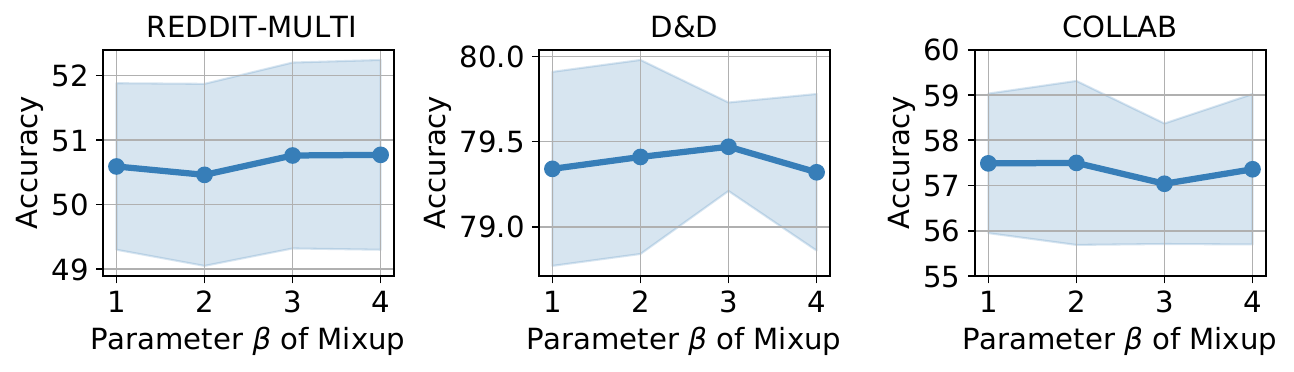}
    \caption{Hyperparamter study on $\beta$ in virtual sample generation.}
    \label{fig:hyper_beta}
\end{figure}

First, when we use manifold mixup to generate virtual training samples, $\lambda$ is randomly sampled from a \texttt{Beta}$(\alpha,\beta)$ distribution. We fix $\alpha=2$ to select the value of $\beta$ from \{1,2,3,4\}.
Figure \ref{fig:hyper_beta} demonstrates that $\beta$ is not sensitive and have little impact on the overall OOD generalization performance. Through our experiment, it is generally recommended to set this value as 2 or 3.

\begin{figure}[h]
    \centering
    \includegraphics[width=\linewidth]{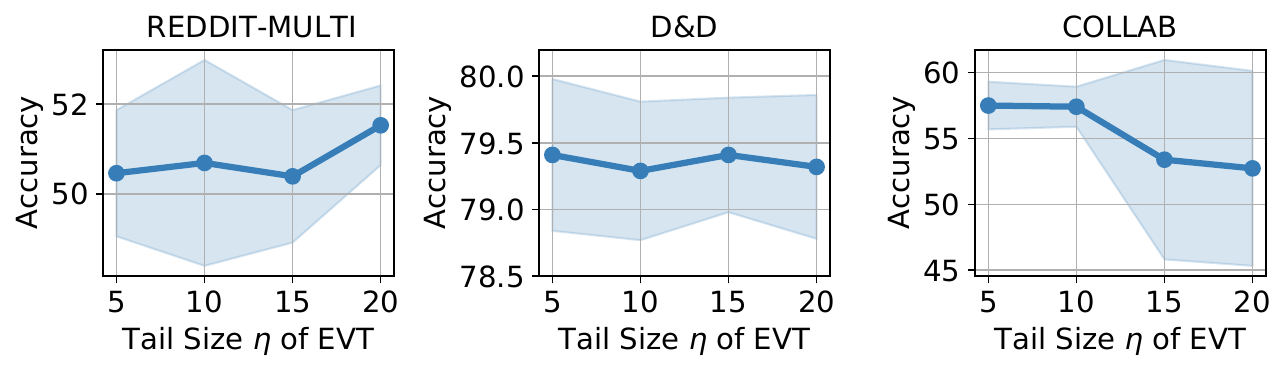}
    \caption{Hyperparamter study on tail size $\tau$ in OOD calibration.}
    \label{fig:hyper_tail}
\end{figure}

Additionally, we conduct a hyperparameter study on the tail size $\eta$ of EVT as shown in Figure \ref{fig:hyper_tail}. Due to various data distributions, the choice of tail length varies. 
For the D\&D dataset, tail size $\eta$ is not sensitive, and the optimal result is reached when tail is equal to 5. However, REDDIT-MULTI and COLLAB dataset is sensitive to this parameter. 
A suitable tail size setting can assist in determining the deviation of manifold mixup samples from the in-distribution data. 

\begin{figure}[h]
    \centering
    \includegraphics[width=\linewidth]{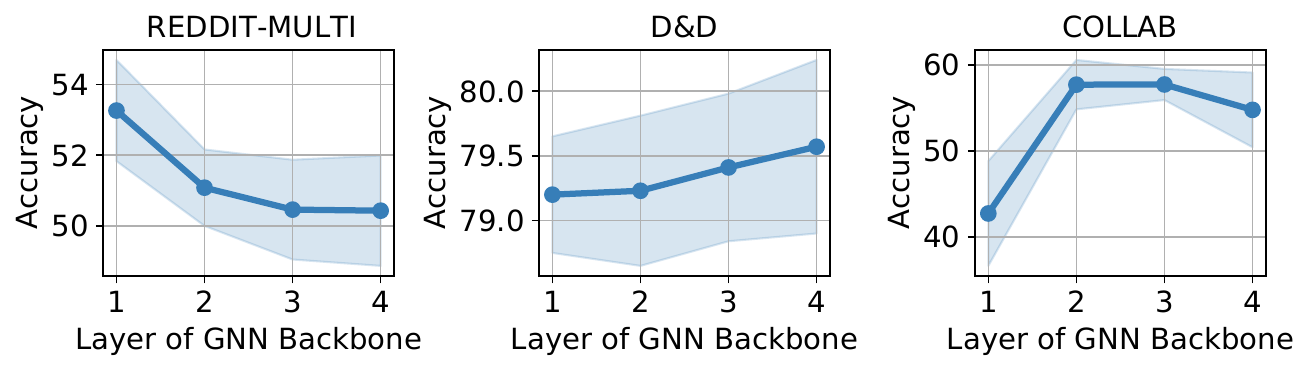}
    \caption{Hyperparamter study on layer of GNN backbone.}
    \label{fig:layer}
\end{figure}

Finally, we investigate the impact of GNN backbone layer, as shown in Figure \ref{fig:layer}. 
Due to the size and density of different graphs, the impact of number of GNN layers is different. The REDDIT-MULTI dataset performs best at one layer, while D\&D and COLLAB perform best at two layers and three layers, respectively.

\section{Conclusion}
\label{sec:conclusion}

In this paper, we propose \texttt{OOD-GMixup} to improve graph OOD generalization by manipulating the training distribution with controllable data augmentation.
We propose to extract the graph rationales and generate virtual training samples with manifold mixup, thereby eliminating the selection bias and structure shortcuts of graphs. 
Through sample reweighting based on OOD confidence score, the representation deviation between ID and OOD distributions is gradually reduced.
Extensive experiments on 6 real-world datasets demonstrate the superiority of our proposed method. 
In the future, we will continue to investigate graph domain generalization with hybrid graph distribution deviations, time-evolving patterns, etc.

\section*{Acknowledgments}
This work was partially supported by National Key R\&D Program of China (No.2022YFB3904204), NSF China (No. 42050105, 62272301, 62020106005, 62061146002, 61960206002).

\bibliographystyle{IEEEtran}
\bibliography{reference}

\begin{thebibliography}{10}
\providecommand{\url}[1]{#1}
\csname url@samestyle\endcsname
\providecommand{\newblock}{\relax}
\providecommand{\bibinfo}[2]{#2}
\providecommand{\BIBentrySTDinterwordspacing}{\spaceskip=0pt\relax}
\providecommand{\BIBentryALTinterwordstretchfactor}{4}
\providecommand{\BIBentryALTinterwordspacing}{\spaceskip=\fontdimen2\font plus
\BIBentryALTinterwordstretchfactor\fontdimen3\font minus
  \fontdimen4\font\relax}
\providecommand{\BIBforeignlanguage}[2]{{%
\expandafter\ifx\csname l@#1\endcsname\relax
\typeout{** WARNING: IEEEtran.bst: No hyphenation pattern has been}%
\typeout{** loaded for the language `#1'. Using the pattern for}%
\typeout{** the default language instead.}%
\else
\language=\csname l@#1\endcsname
\fi
#2}}
\providecommand{\BIBdecl}{\relax}
\BIBdecl

\bibitem{DBLP:conf/iclr/SunHV020}
F.~Sun, J.~Hoffmann, V.~Verma, and J.~Tang, ``Infograph: Unsupervised and
  semi-supervised graph-level representation learning via mutual information
  maximization,'' in \emph{8th International Conference on Learning
  Representations, {ICLR} 2020, Addis Ababa, Ethiopia, April 26-30, 2020},
  2020.

\bibitem{DBLP:conf/aaai/Hassani22}
K.~Hassani, ``Cross-domain few-shot graph classification,'' in
  \emph{Thirty-Sixth {AAAI} Conference on Artificial Intelligence, {AAAI} 2022,
  February 22 - March 1, 2022}, 2022, pp. 6856--6864.

\bibitem{DBLP:conf/ijcai/WuLL0022}
J.~Wu, S.~Li, J.~Li, Y.~Pan, and K.~Xu, ``A simple yet effective method for
  graph classification,'' in \emph{Proceedings of the Thirty-First
  International Joint Conference on Artificial Intelligence, {IJCAI} 2022,
  Vienna, Austria, 23-29 July 2022}.\hskip 1em plus 0.5em minus 0.4em\relax
  ijcai.org, 2022, pp. 3580--3586.

\bibitem{OOD-GNN}
H.~Li, X.~Wang, Z.~Zhang, and W.~Zhu, ``Ood-gnn: Out-of-distribution
  generalized graph neural network,'' \emph{IEEE Transactions on Knowledge and
  Data Engineering}, pp. 1--14, 2022.

\bibitem{GNN-DVD}
S.~Fan, X.~Wang, C.~Shi, K.~Kuang, N.~Liu, and B.~Wang, ``Debiased graph neural
  networks with agnostic label selection bias,'' \emph{IEEE Transactions on
  Neural Networks and Learning Systems}, pp. 1--12, 2022.

\bibitem{DIR-GNN}
\BIBentryALTinterwordspacing
Y.~Wu, X.~Wang, A.~Zhang, X.~He, and T.-S. Chua, ``Discovering invariant
  rationales for graph neural networks,'' in \emph{International Conference on
  Learning Representations}, 2022. [Online]. Available:
  \url{https://openreview.net/forum?id=hGXij5rfiHw}
\BIBentrySTDinterwordspacing

\bibitem{chen2022invariance}
Y.~Chen, Y.~Zhang, H.~Yang, K.~Ma, B.~Xie, T.~Liu, B.~Han, and J.~Cheng,
  ``Invariance principle meets out-of-distribution generalization on graphs,''
  in \emph{ICML Workshop on Spurious Correlations, Invariance and Stability},
  2022.

\bibitem{buffelli2022sizeshiftreg}
\BIBentryALTinterwordspacing
D.~Buffelli, P.~Lio, and F.~Vandin, ``Sizeshiftreg: a regularization method for
  improving size-generalization in graph neural networks,'' in \emph{Advances
  in Neural Information Processing Systems}, A.~H. Oh, A.~Agarwal, D.~Belgrave,
  and K.~Cho, Eds., 2022. [Online]. Available:
  \url{https://openreview.net/forum?id=wOI0AUAq9BR}
\BIBentrySTDinterwordspacing

\bibitem{virtual_node}
J.~Gilmer, S.~S. Schoenholz, P.~F. Riley, O.~Vinyals, and G.~E. Dahl, ``Neural
  message passing for quantum chemistry,'' in \emph{Proceedings of the 34th
  International Conference on Machine Learning, {ICML} 2017, Sydney, NSW,
  Australia, 6-11 August 2017}, vol.~70, 2017, pp. 1263--1272.

\bibitem{dropedge}
Y.~Rong, W.~Huang, T.~Xu, and J.~Huang, ``Dropedge: Towards deep graph
  convolutional networks on node classification,'' in \emph{8th International
  Conference on Learning Representations, {ICLR} 2020, Addis Ababa, Ethiopia,
  April 26-30, 2020}, 2020.

\bibitem{Dropnode}
P.~A. Papp, K.~Martinkus, L.~Faber, and R.~Wattenhofer, ``Dropgnn: Random
  dropouts increase the expressiveness of graph neural networks,'' in
  \emph{Advances in Neural Information Processing Systems, NeurIPS 2021,
  December 6-14, 2021, virtual}, 2021, pp. 21\,997--22\,009.

\bibitem{DBLP:conf/www/WangWLCH21}
Y.~Wang, W.~Wang, Y.~Liang, Y.~Cai, and B.~Hooi, ``Mixup for node and graph
  classification,'' in \emph{{WWW} '21: The Web Conference 2021, Virtual Event
  / Ljubljana, Slovenia, April 19-23, 2021}, 2021, pp. 3663--3674.

\bibitem{flag}
K.~Kong, G.~Li, M.~Ding, Z.~Wu, C.~Zhu, B.~Ghanem, G.~Taylor, and T.~Goldstein,
  ``Robust optimization as data augmentation for large-scale graphs,'' in
  \emph{{IEEE} Conference on Computer Vision and Pattern Recognition, {CVPR}
  2022, New Orleans, Louisiana, June 19-24, 2022}, 2022.

\bibitem{DisenGCN}
J.~Ma, P.~Cui, K.~Kuang, X.~Wang, and W.~Zhu, ``Disentangled graph
  convolutional networks,'' in \emph{Proceedings of the 36th International
  Conference on Machine Learning, {ICML} 2019, 9-15 June 2019, Long Beach,
  California, {USA}}, vol.~97, 2019, pp. 4212--4221.

\bibitem{FactorGCN}
Y.~Yang, Z.~Feng, M.~Song, and X.~Wang, ``Factorizable graph convolutional
  networks,'' in \emph{Advances in Neural Information Processing Systems,
  NeurIPS 2020, December 6-12, 2020, virtual}, 2020.

\bibitem{TopK}
H.~Gao and S.~Ji, ``Graph u-nets,'' in \emph{Proceedings of the 36th
  International Conference on Machine Learning, {ICML} 2019, 9-15 June 2019,
  Long Beach, California, {USA}}, vol.~97, 2019, pp. 2083--2092.

\bibitem{SAG}
J.~Lee, I.~Lee, and J.~Kang, ``Self-attention graph pooling,'' in
  \emph{Proceedings of the 36th International Conference on Machine Learning,
  {ICML} 2019, 9-15 June 2019, Long Beach, California, {USA}}, vol.~97, 2019,
  pp. 3734--3743.

\bibitem{PNA}
G.~Corso, L.~Cavalleri, D.~Beaini, P.~Li{\`{o}}, and P.~Velickovic, ``Principal
  neighbourhood aggregation for graph nets,'' in \emph{Advances in Neural
  Information Processing Systems, NeurIPS 2020, December 6-12, 2020, virtual},
  2020.

\bibitem{IRM}
M.~Arjovsky, L.~Bottou, I.~Gulrajani, and D.~Lopez{-}Paz, ``Invariant risk
  minimization,'' \emph{CoRR}, vol. 1907.02893, 2019.

\bibitem{GroupDRO}
S.~Sagawa, P.~W. Koh, T.~B. Hashimoto, and P.~Liang, ``Distributionally robust
  neural networks,'' in \emph{8th International Conference on Learning
  Representations, {ICLR} 2020, Addis Ababa, Ethiopia, April 26-30, 2020},
  2020.

\bibitem{StableNet}
X.~Zhang, P.~Cui, R.~Xu, L.~Zhou, Y.~He, and Z.~Shen, ``Deep stable learning
  for out-of-distribution generalization,'' in \emph{{IEEE} Conference on
  Computer Vision and Pattern Recognition, {CVPR} 2021, virtual, June 19-25,
  2021}, 2021, pp. 5372--5382.

\bibitem{DBLP:conf/icml/WuZVR20}
S.~Wu, H.~R. Zhang, G.~Valiant, and C.~R{\'{e}}, ``On the generalization
  effects of linear transformations in data augmentation,'' in
  \emph{Proceedings of the 37th International Conference on Machine Learning,
  {ICML} 2020, 13-18 July 2020, Virtual Event}, vol. 119.\hskip 1em plus 0.5em
  minus 0.4em\relax {PMLR}, 2020, pp. 10\,410--10\,420.

\bibitem{DBLP:conf/iclr/ZhangDKG021}
L.~Zhang, Z.~Deng, K.~Kawaguchi, A.~Ghorbani, and J.~Zou, ``How does mixup help
  with robustness and generalization?'' in \emph{9th International Conference
  on Learning Representations, {ICLR} 2021, Virtual Event, Austria, May 3-7,
  2021}, 2021.

\bibitem{DBLP:conf/aaai/MoPXS022}
Y.~Mo, L.~Peng, J.~Xu, X.~Shi, and X.~Zhu, ``Simple unsupervised graph
  representation learning,'' in \emph{Thirty-Sixth {AAAI} Conference on
  Artificial Intelligence, {AAAI} 2022, Thirty-Fourth Conference on Innovative
  Applications of Artificial Intelligence, {IAAI} 2022, The Twelveth Symposium
  on Educational Advances in Artificial Intelligence, {EAAI} 2022 Virtual
  Event, February 22 - March 1, 2022}.\hskip 1em plus 0.5em minus 0.4em\relax
  {AAAI} Press, 2022, pp. 7797--7805.

\bibitem{DBLP:conf/nips/YouCSCWS20}
Y.~You, T.~Chen, Y.~Sui, T.~Chen, Z.~Wang, and Y.~Shen, ``Graph contrastive
  learning with augmentations,'' in \emph{Advances in Neural Information
  Processing Systems 33: Annual Conference on Neural Information Processing
  Systems 2020, NeurIPS 2020, December 6-12, 2020, virtual}, 2020.

\bibitem{DBLP:journals/sigkdd/DingXTL22}
K.~Ding, Z.~Xu, H.~Tong, and H.~Liu, ``Data augmentation for deep graph
  learning: {A} survey,'' \emph{{SIGKDD} Explor.}, vol.~24, no.~2, pp. 61--77,
  2022.

\bibitem{de2006extreme}
L.~De~Haan, A.~Ferreira, and A.~Ferreira, \emph{Extreme value theory: an
  introduction}.\hskip 1em plus 0.5em minus 0.4em\relax Springer, 2006,
  vol.~21.

\bibitem{einmahl2011ultimate}
J.~H. Einmahl and S.~G. Smeets, ``Ultimate 100-m world records through
  extreme-value theory,'' \emph{Statistica Neerlandica}, vol.~65, no.~1, pp.
  32--42, 2011.

\bibitem{tippett2016more}
M.~K. Tippett, C.~Lepore, and J.~E. Cohen, ``More tornadoes in the most extreme
  us tornado outbreaks,'' \emph{Science}, vol. 354, no. 6318, pp. 1419--1423,
  2016.

\bibitem{songchitruksa2006extreme}
P.~Songchitruksa and A.~P. Tarko, ``The extreme value theory approach to safety
  estimation,'' \emph{Accident Analysis \& Prevention}, vol.~38, no.~4, pp.
  811--822, 2006.

\bibitem{robustness-analysis}
T.~Weng, H.~Zhang, P.~Chen, J.~Yi, D.~Su, Y.~Gao, C.~Hsieh, and L.~Daniel,
  ``Evaluating the robustness of neural networks: An extreme value theory
  approach,'' in \emph{6th International Conference on Learning
  Representations, {ICLR} 2018, Vancouver, BC, Canada, April 30 - May 3, 2018},
  2018.

\bibitem{clustering}
S.~Zheng, K.~Fan, Y.~Hou, J.~Feng, and Y.~Fu, ``Clustering by the probability
  distributions from extreme value theory,'' \emph{IEEE Transactions on
  Artificial Intelligence}, pp. 1--1, 2022.

\bibitem{PMOSR}
W.~J. Scheirer, L.~P. Jain, and T.~E. Boult, ``Probability models for open set
  recognition,'' \emph{{IEEE} Trans. Pattern Anal. Mach. Intell.}, vol.~36,
  no.~11, pp. 2317--2324, 2014.

\bibitem{Openmax}
A.~Bendale and T.~E. Boult, ``Towards open set deep networks,'' in \emph{2016
  {IEEE} Conference on Computer Vision and Pattern Recognition, {CVPR} 2016,
  Las Vegas, NV, USA, June 27-30, 2016}, 2016, pp. 1563--1572.

\bibitem{C2AE}
P.~Oza and V.~M. Patel, ``{C2AE:} class conditioned auto-encoder for open-set
  recognition,'' in \emph{{IEEE} Conference on Computer Vision and Pattern
  Recognition, {CVPR} 2019, Long Beach, CA, USA, June 16-20, 2019}, 2019, pp.
  2307--2316.

\bibitem{libMR}
W.~J. Scheirer, A.~Rocha, R.~J. Micheals, and T.~E. Boult, ``Meta-recognition:
  The theory and practice of recognition score analysis,'' \emph{{IEEE} Trans.
  Pattern Anal. Mach. Intell.}, vol.~33, no.~8, pp. 1689--1695, 2011.

\bibitem{mixup}
H.~Zhang, M.~Ciss{\'{e}}, Y.~N. Dauphin, and D.~Lopez{-}Paz, ``mixup: Beyond
  empirical risk minimization,'' in \emph{6th International Conference on
  Learning Representations, {ICLR} 2018, Vancouver, BC, Canada, April 30 - May
  3, 2018}, 2018.

\bibitem{DBLP:conf/icml/VermaLBNMLB19}
V.~Verma, A.~Lamb, C.~Beckham, A.~Najafi, I.~Mitliagkas, D.~Lopez{-}Paz, and
  Y.~Bengio, ``Manifold mixup: Better representations by interpolating hidden
  states,'' in \emph{Proceedings of the 36th International Conference on
  Machine Learning, {ICML} 2019, 9-15 June 2019, Long Beach, California,
  {USA}}, vol.~97, 2019, pp. 6438--6447.

\bibitem{DBLP:conf/kdd/SuiWWL0C22}
Y.~Sui, X.~Wang, J.~Wu, M.~Lin, X.~He, and T.~Chua, ``Causal attention for
  interpretable and generalizable graph classification,'' in \emph{28th {ACM}
  {SIGKDD} Conference on Knowledge Discovery and Data Mining, Washington, DC,
  USA, August 14 - 18, 2022}.\hskip 1em plus 0.5em minus 0.4em\relax {ACM},
  2022, pp. 1696--1705.

\bibitem{DBLP:journals/corr/abs-2204-11028}
X.~Wang, Y.~Wu, A.~Zhang, F.~Feng, X.~He, and T.~Chua, ``Reinforced causal
  explainer for graph neural networks,'' \emph{{IEEE} Trans. Pattern Anal.
  Mach. Intell.}, 2022.

\bibitem{DBLP:conf/nips/SnellSZ17}
J.~Snell, K.~Swersky, and R.~S. Zemel, ``Prototypical networks for few-shot
  learning,'' in \emph{Advances in Neural Information Processing Systems, NIPS
  2017, December 4-9, 2017, Long Beach, CA, {USA}}, 2017, pp. 4077--4087.

\bibitem{DBLP:conf/kdd/Geometer}
B.~Lu, X.~Gan, L.~Yang, W.~Zhang, L.~Fu, and X.~Wang, ``Geometer: Graph
  few-shot class-incremental learning via prototype representation,'' in
  \emph{The 28th {ACM} SIGKDD Conference on Knowledge Discovery and Data
  Mining, Washington, DC, USA, August 14--18, 2022}, 2022.

\bibitem{vrm}
O.~Chapelle, J.~Weston, L.~Bottou, and V.~Vapnik, ``Vicinal risk
  minimization,'' in \emph{Advances in Neural Information Processing Systems,
  NIPS 2000, Denver, CO, {USA}}, 2000, pp. 416--422.

\bibitem{GAT}
P.~Velickovic, G.~Cucurull, A.~Casanova, A.~Romero, P.~Li{\`{o}}, and
  Y.~Bengio, ``Graph attention networks,'' in \emph{6th International
  Conference on Learning Representations, {ICLR} 2018, Vancouver, BC, Canada,
  April 30 - May 3, 2018}, 2018.

\bibitem{vrex}
D.~Krueger, E.~Caballero, J.~Jacobsen, A.~Zhang, J.~Binas, D.~Zhang, R.~L.
  Priol, and A.~C. Courville, ``Out-of-distribution generalization via risk
  extrapolation (rex),'' in \emph{Proceedings of the 38th International
  Conference on Machine Learning, {ICML} 2021, 18-24 July 2021, Virtual Event},
  vol. 139, 2021, pp. 5815--5826.

\bibitem{ib-irm}
K.~Ahuja, E.~Caballero, D.~Zhang, J.~Gagnon{-}Audet, Y.~Bengio, I.~Mitliagkas,
  and I.~Rish, ``Invariance principle meets information bottleneck for
  out-of-distribution generalization,'' in \emph{Advances in Neural Information
  Processing System, NeurIPS 2021, December 6-14, 2021, virtual}, 2021, pp.
  3438--3450.

\bibitem{DBLP:conf/iclr/RuanDM22}
Y.~Ruan, Y.~Dubois, and C.~J. Maddison, ``Optimal representations for covariate
  shift,'' in \emph{The Tenth International Conference on Learning
  Representations, {ICLR} 2022, Virtual Event, April 25-29, 2022}, 2022.

\bibitem{wu2022handling}
Q.~Wu, H.~Zhang, J.~Yan, and D.~Wipf, ``Handling distribution shifts on graphs:
  An invariance perspective,'' in \emph{International Conference on Learning
  Representations}, 2022.

\bibitem{G-Mixup}
X.~Han, Z.~Jiang, N.~Liu, and X.~Hu, ``G-mixup: Graph data augmentation for
  graph classification,'' in \emph{International Conference on Machine
  Learning, {ICML} 2022, 17-23 July 2022, Baltimore, Maryland, {USA}}, vol.
  162, 2022, pp. 8230--8248.

\bibitem{focal_loss}
T.~Lin, P.~Goyal, R.~B. Girshick, K.~He, and P.~Doll{\'{a}}r, ``Focal loss for
  dense object detection,'' in \emph{{IEEE} International Conference on
  Computer Vision, {ICCV} 2017, Venice, Italy, October 22-29, 2017}, 2017, pp.
  2999--3007.

\end{thebibliography}

\newpage

\section{Biography Section}
 

\begin{IEEEbiography}[{\includegraphics[width=1in,height=1.25in,clip,keepaspectratio]{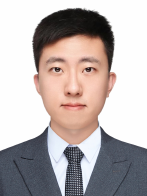}}]{Bin Lu}
received the B.E. degree in information engineering from Shanghai Jiao Tong University, Shanghai, China, in 2020. He is currently working toward the PhD degree in the Department of Electronic Engineering, Shanghai Jiao Tong University. His research interests are in the area of graph neural network, spatiotemporal data mining, AI for Science.
\end{IEEEbiography}


\begin{IEEEbiography}[{\includegraphics[width=1in,height=1.25in,clip,keepaspectratio]{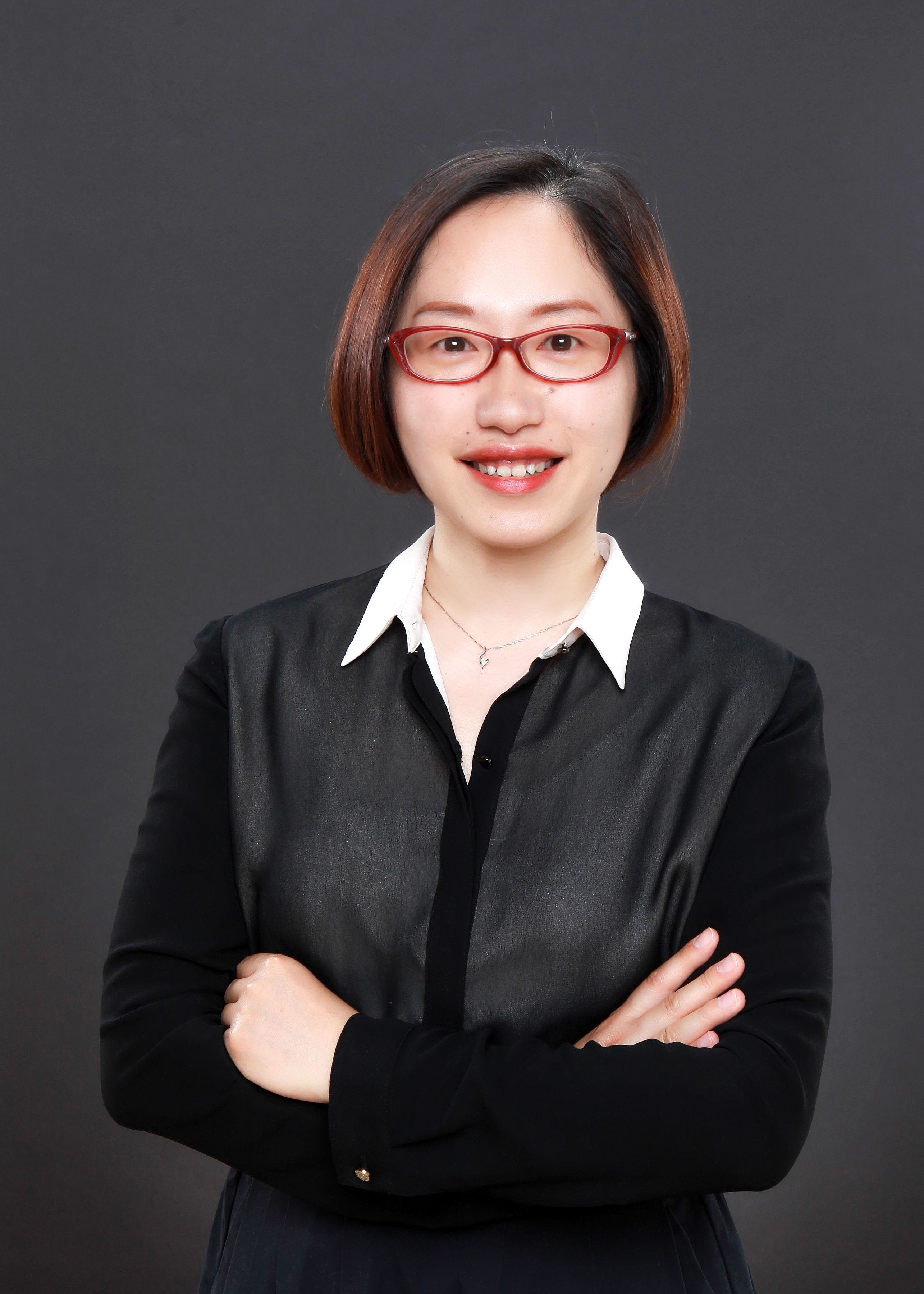}}]{Xiaoying Gan}
received the PhD degree in electronic engineering from the Shanghai Jiao Tong University, Shanghai, China, in 2006. From 2009 to 2010, she was a visiting researcher with the California Institute for Telecommunications and Information Technology (Calit2), University of California San Diego, La Jolla, California. She is currently a professor at the Department of Electronic Engineering, Shanghai Jiao Tong University. Her current research interests include crowd computing, data mining, and resource management in Internet of Things.
\end{IEEEbiography}

\begin{IEEEbiography}[{\includegraphics[width=1in,height=1.25in,clip,keepaspectratio]{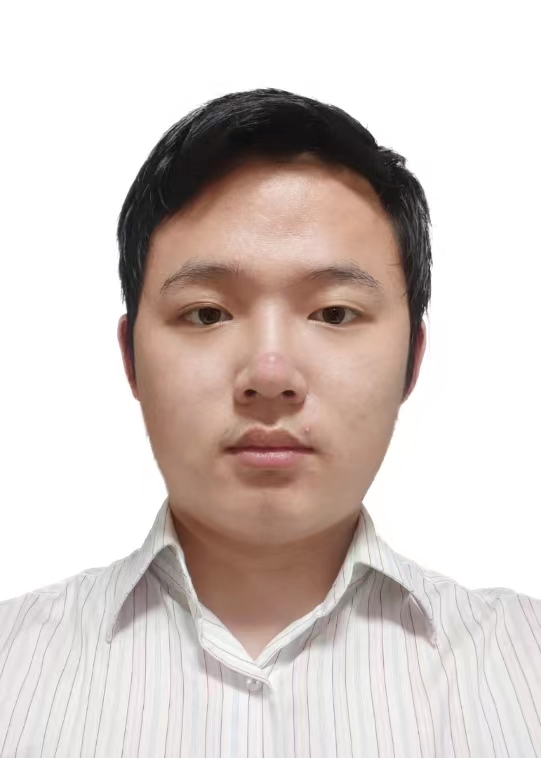}}]{Ze Zhao}
is currently working toward the B.E. degree in the Department of Electronic Engineering, Shanghai Jiao Tong University, Shanghai, China. He is currently working as a research intern supervised by Prof. Xiaoying Gan. His research interests include graph neural network, spatiotemporal data mining, knowledge graph.

\end{IEEEbiography}


\begin{IEEEbiography}[{\includegraphics[width=1in,height=1.25in,clip,keepaspectratio]{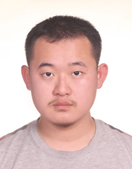}}]{Shiyu Liang}
received the B.E degree in electronic engineering from the Shanghai Jiao Tong University, Shanghai, China, in 2015, the PhD degree in Electrical and Computer Engineering at University of Illinois at Urbana-Champaign, in 2021. He is currently an assistant professor with John Hopcroft Center for Computer Science at Shanghai Jiao Tong University. His research of interests include deep learning, machine learning, optimization and applied probability.
\end{IEEEbiography}


\begin{IEEEbiography}[{\includegraphics[width=1in,height=1.25in,clip,keepaspectratio]{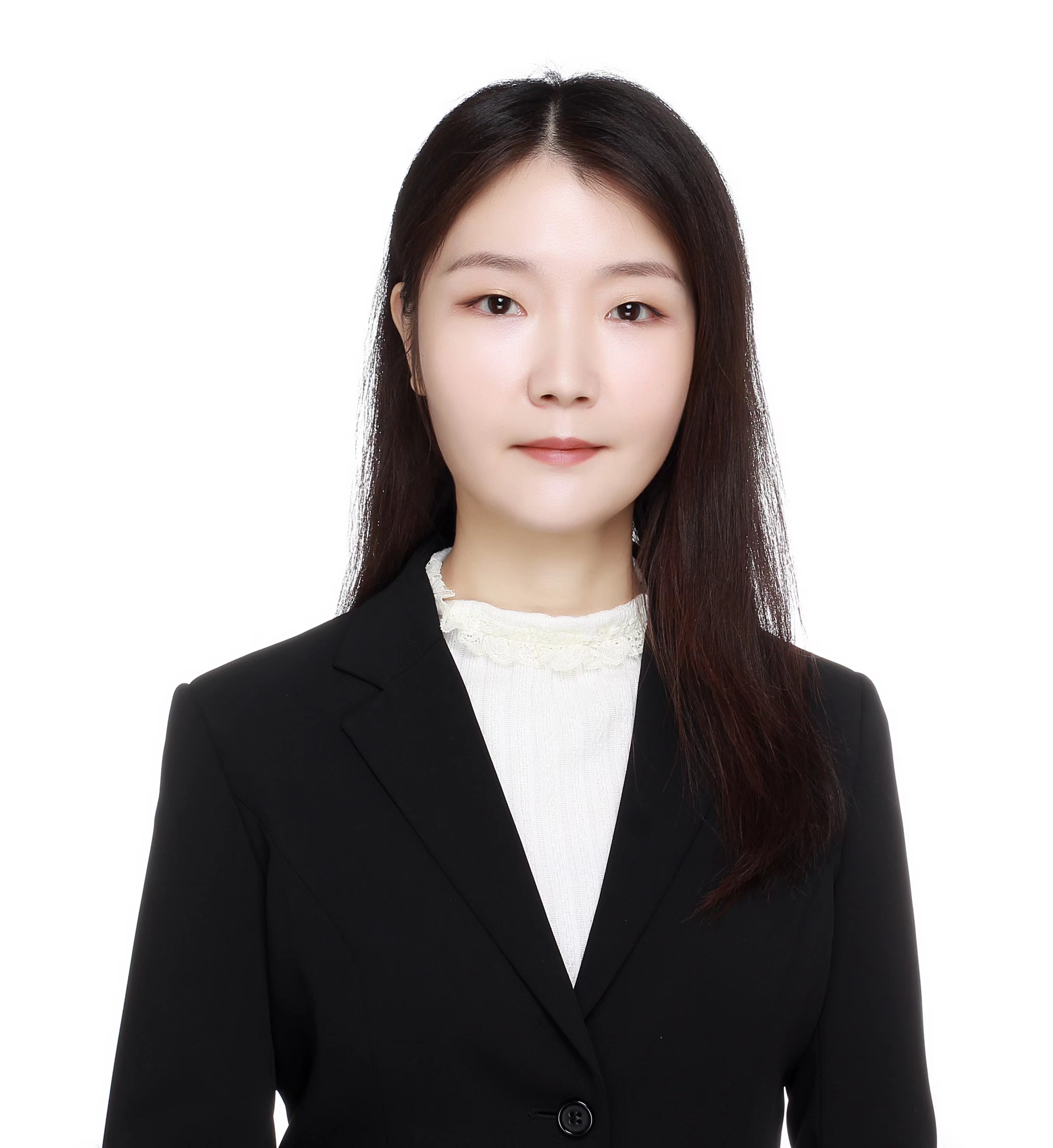}}]{Luoyi Fu}
received the B.E. degree in electronic engineering and the PhD degree in computer science and engineering from Shanghai Jiao Tong University, China, in 2009 and 2015, respectively. She is currently an associate professor with the Department of Computer Science and Engineering, Shanghai Jiao Tong University. Her research of interests include social networking and big data, connectivity analysis, and random graphs. She has been an editor of IEEE/ACM Transactions on Networking and a member of the Technical Program Committees of several conferences, including ACM MobiHoc from 2018 to 2022 and IEEE INFOCOM from 2018 to 2022.
\end{IEEEbiography}

\begin{IEEEbiography}[{\includegraphics[width=1in,height=1.25in,clip,keepaspectratio]{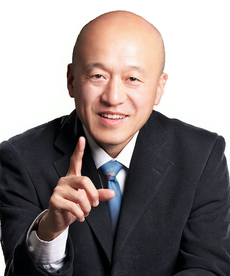}}]{Xinbing Wang}
received the BS degree (Hons.) from the Department of Automation, Shanghai Jiao Tong University, Shanghai, China, in 1998, the MS degree from the Department of Computer Science and Technology, Tsinghua University, Beijing, China, in 2001, and the PhD degree, majoring in the electrical and computer engineering and minoring in mathematics, from North Carolina State University, Raleigh, North Carolina, in 2006. He is currently a distinguished professor with the Department of Electronic Engineering, Shanghai Jiao Tong University, China. He has been an Editor at Large for the IEEE/ACM Transactions on Networking, and the Associate Editor for IEEE Transactions on Information Theory, and a member of the Technical Program Committees of several conferences including IEEE INFOCOM 2009$-$-2023.
\end{IEEEbiography}

\begin{IEEEbiography}[{\includegraphics[width=1in,height=1.25in,clip,keepaspectratio]{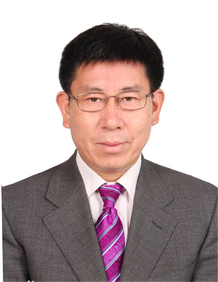}}]{Chenghu Zhou}
received the BS degree in geography from Nanjing University, Nanjing, China, in 1984, and the MS and PhD degrees in geographic information system from the Chinese Academy of Sciences (CAS), Beijing, China, in 1987 and 1992, respectively. He is currently an academician with the Chinese Academy of Sciences, China, where he is also a research professor with the Institute of Geographical Sciences and Natural Resources Research, and a professor with the School of Geography and Ocean Science, Nanjing University, China. His research interests include spatial and temporal data mining, geographic modeling, hydrology and water resources, and geographic information systems and remote sensing applications.
\end{IEEEbiography}

\vfill

\end{document}